\documentclass[table]{ai2style/ai2}

\usepackage{microtype}
\usepackage{hyperref}
\usepackage{url}
\usepackage{booktabs}
\usepackage{graphicx}
\usepackage{lineno}
\usepackage{enumitem}
\usepackage{listings} 
\usepackage{svg}

\definecolor{ darkblue}{rgb}{0, 0, 0.5}
\hypersetup{colorlinks=true, citecolor=darkblue, linkcolor=darkblue, urlcolor=darkblue}

\usepackage{amssymb}
\usepackage{multirow}
\usepackage{bigdelim}
\usepackage{todonotes}
\usepackage{longtable}
\usepackage{tabularray}
\usepackage{wrapfig}
\usepackage[most]{tcolorbox}
\usepackage{url}
\usepackage{xspace}
\usepackage{svg}
\usepackage[absolute]{textpos} 

\usepackage{fdsymbol}   

\usepackage[utf8]{inputenc} 
\usepackage[T1]{fontenc}    
\usepackage{url}            
\usepackage{booktabs}       
\usepackage{amsfonts}       
\usepackage{nicefrac}       
\usepackage{microtype}      
\usepackage[table]{xcolor}         
\usepackage{amsmath}
\usepackage[most]{tcolorbox}
\usepackage{csquotes}

\usepackage{siunitx}
\usepackage{graphicx}
\usepackage{arydshln}
\usepackage{wrapfig}
\usepackage{enumitem}
\usepackage{soul} 

\usepackage{multirow}
\usepackage{xspace}
\usepackage{adjustbox}
\usepackage{pifont}
\usepackage{caption}
\usepackage{makecell}
\usepackage{subcaption}
\usepackage{bold-extra}
\usepackage{url}

\usepackage{pgf-pie}

\usepackage{hyperref}
\definecolor{linkcolor}{RGB}{0, 0, 128}
\hypersetup{
     colorlinks   = true,
     citecolor    = linkcolor,
     linkcolor    = linkcolor,
     urlcolor     = linkcolor,
}
\usepackage{pifont}
\usepackage{listings}

\setlist[itemize]{leftmargin=*,itemsep=0em,parsep=0.3em,topsep=0.3em}

\DeclareUnicodeCharacter{2212}{\ensuremath{-}}

\definecolor{maroon}{HTML}{F26035}
\definecolor{yellow}{HTML}{FDBC42}
\definecolor{lavender}{HTML}{734f96}
\definecolor{darkergrey}{HTML}{444444}
\definecolor{midgrey}{HTML}{e6eded}
\definecolor{ai2pink}{HTML}{f0529c}
\definecolor{ai2midpink}{HTML}{fad3e5}
\definecolor{ai2lightpink}{HTML}{fbecf3}
\definecolor{ai2midwhite}{HTML}{f2e5d9}
\definecolor{ai2offwhite}{HTML}{fbf4ee}
\definecolor{ai2green}{HTML}{0fcb8c}
\definecolor{ai2lightgreen}{HTML}{e7f9f3}
\definecolor{ai2darkgreen}{HTML}{105257}
\definecolor{ai2purple}{HTML}{B932EB}
\definecolor{ai2lightpurple}{HTML}{f7e8fc}
\definecolor{neutralEight}{HTML}{343434}
\definecolor{neutralFive}{HTML}{838383}
\definecolor{neutralThree}{HTML}{bebebe}
\definecolor{neutralOne}{HTML}{dedede}
\definecolor{lightgrey}{HTML}{fafcfc}

\usepackage{tikz}

\definecolor{maroon}{HTML}{F26035}
\definecolor{yellow}{HTML}{FDBC42}
\definecolor{darkred}{RGB}{156, 39, 33}
\definecolor{darkblue}{RGB}{31, 90, 153}
\definecolor{forestgreen}{rgb}{0.13, 0.55, 0.13}
\definecolor{olmoDarkBlue}{HTML}{012e59}
\definecolor{olmoBlue}{HTML}{265ed4}
\definecolor{olmoLightBlue}{HTML}{012e59}
\definecolor{olmoTeal}{HTML}{00d5ff}
\definecolor{olmoYellow}{HTML}{ffbb00}
\definecolor{olmoOrange}{HTML}{ff9100}

\usepackage{setspace}

\usepackage{nicematrix}
\newcolumntype{L}[1]{>{\raggedright\let\newline\\\arraybackslash\hspace{0pt}}m{#1}}
\newcolumntype{C}[1]{>{\centering\let\newline\\\arraybackslash\hspace{0pt}}m{#1}}
\newcolumntype{R}[1]{>{\raggedleft\let\newline\\\arraybackslash\hspace{0pt}}m{#1}}
\newcolumntype{P}[1]{>{\centering\let\newline\\\arraybackslash\columncolor{ai2lightpink}}m{#1}}
\newcommand{\aitwo}{\raisebox{-1.5pt}{\includegraphics[height=1.05em]{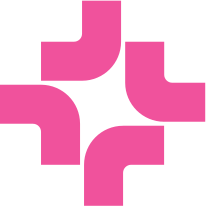}}\xspace}

\title{\pipeline: Unlocking Trillions of Tokens in PDFs with Vision Language Models}

\newcommand{\pipeline}{\textsc{olmOCR}\xspace}
\newcommand{\olmocr}{\pipeline{}\xspace}
\newcommand{\pesto}{\textsc{olmOCR-peS2o}\xspace}
\newcommand{\bench}{\textsc{olmOCR-Bench}\xspace}
\newcommand{\model}{\texttt{olmOCR-7B-0225-preview}\xspace}
\newcommand{\train}{\texttt{olmOCR-mix-0225}\xspace}
\newcommand{\method}{\textsc{document-anchoring}\xspace}
\newcommand{\GptFourO}{GPT-4o\xspace}

\newcommand{\marker}{\textsc{Marker}\xspace}
\newcommand{\gotocr}{\textsc{GOTOCR}\xspace}
\newcommand{\mineru}{\textsc{MinerU}\xspace}

\newcommand{\huggingface}{\raisebox{-1.5pt}{\includegraphics[height=1.05em]{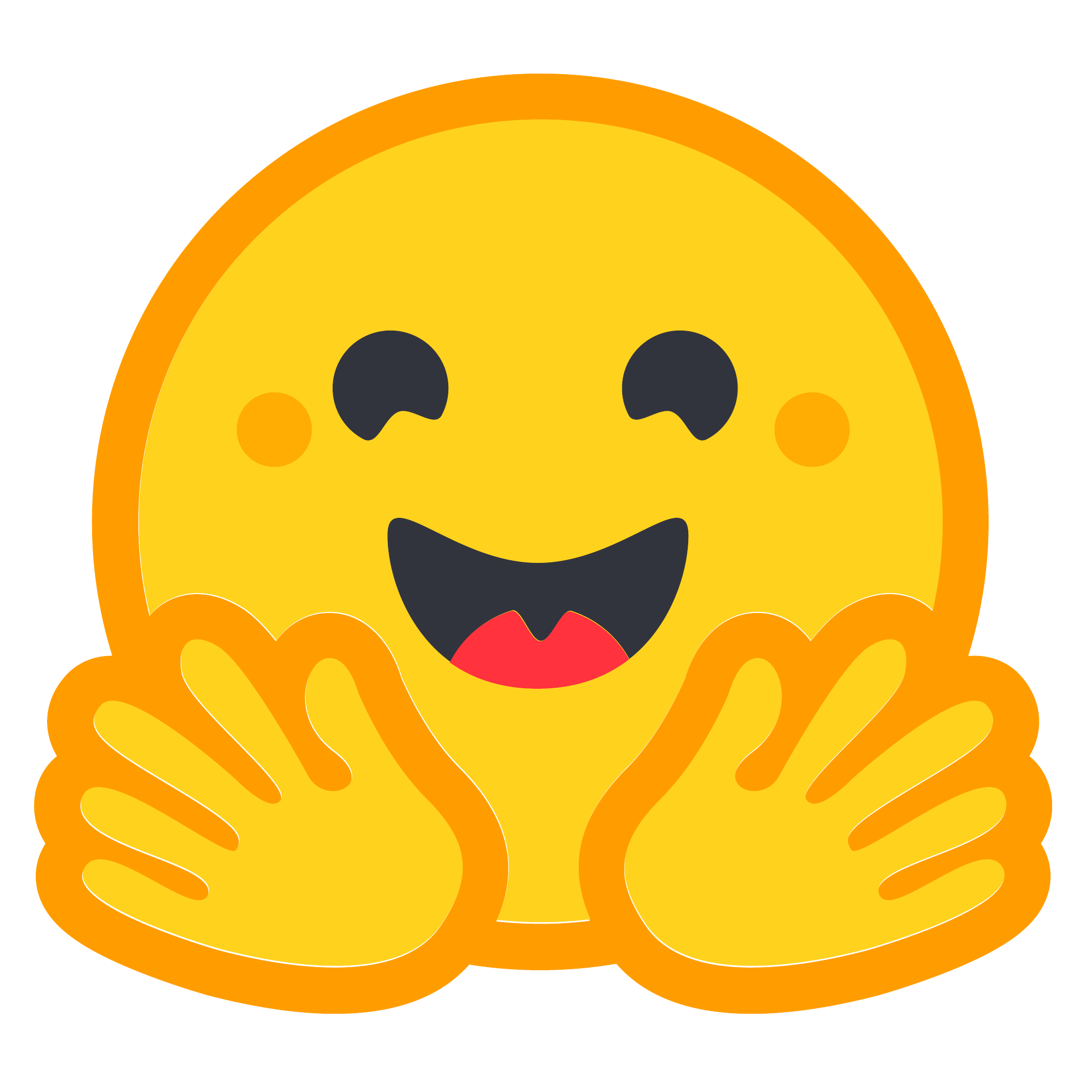}}\xspace}
\newcommand{\coreContrib}{\raisebox{.28em}{\hspace{.05em}\includegraphics[height=.45em]{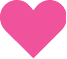}}\xspace}

\newcommand{\github}{\raisebox{-1.5pt}{\includegraphics[height=1.05em]{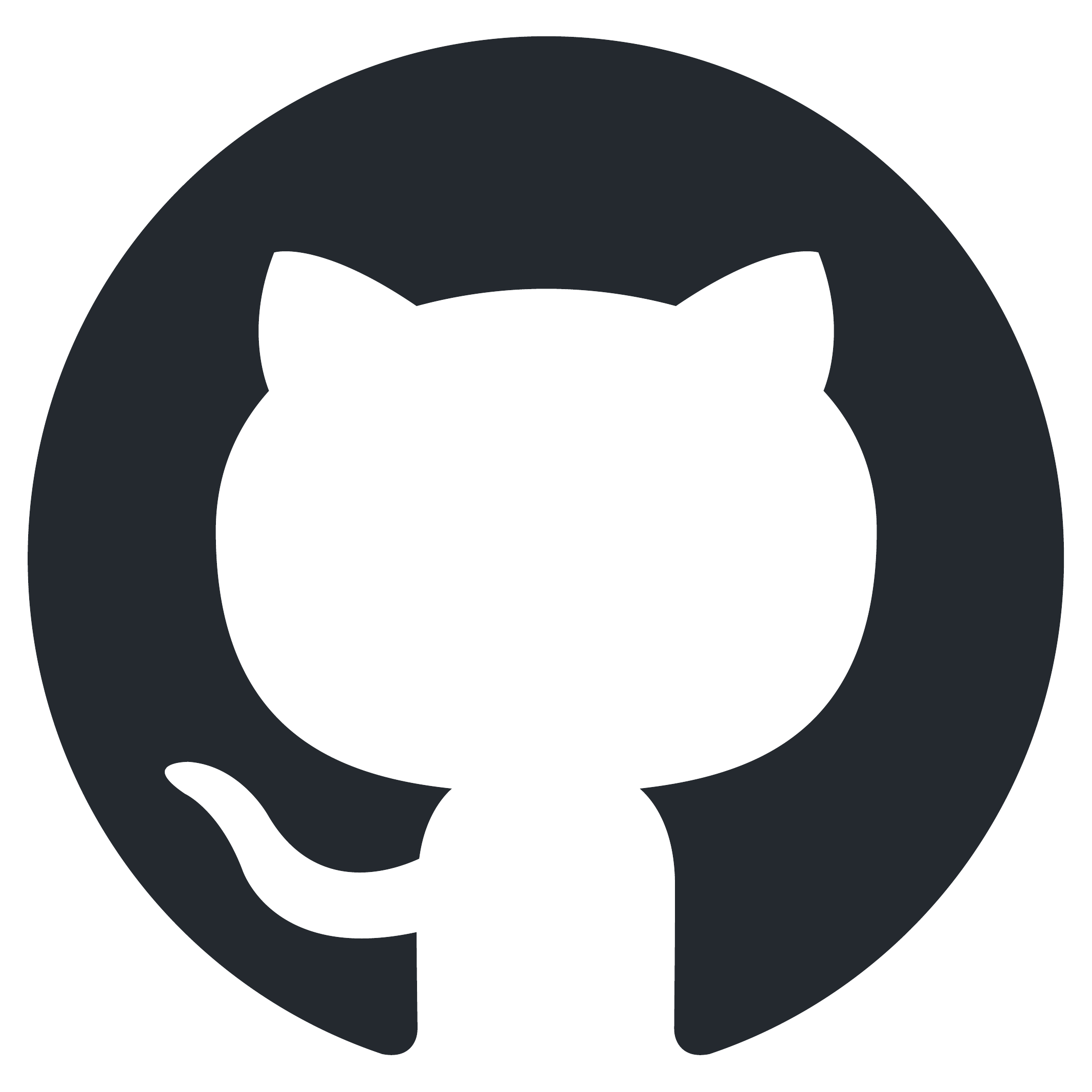}}\xspace}

\authorOne[]{Jake Poznanski\coreContrib}
\authorOne[]{Aman Rangapur}

\authorTwo[]{Jon Borchardt}
\authorTwo[]{Jason Dunkelberger}
\authorTwo[]{Regan Huff}
\authorTwo[]{Daniel Lin}
\authorTwo[]{Christopher Wilhelm}

\authorThree[]{Kyle Lo\coreContrib}
\authorThree[]{Luca Soldaini\coreContrib}

\contribution[]{Allen Institute for AI, Seattle, USA \quad \texttt{\{jakep|kylel|lucas\}@allenai.org} \quad \coreContrib{ indicates core contributors.}}

\setmainfigure{
\vspace{-0.5cm}
\begin{figure}[!h]
    \centering
    \vspace*{\fill}  
    \includegraphics[width=0.75\linewidth]{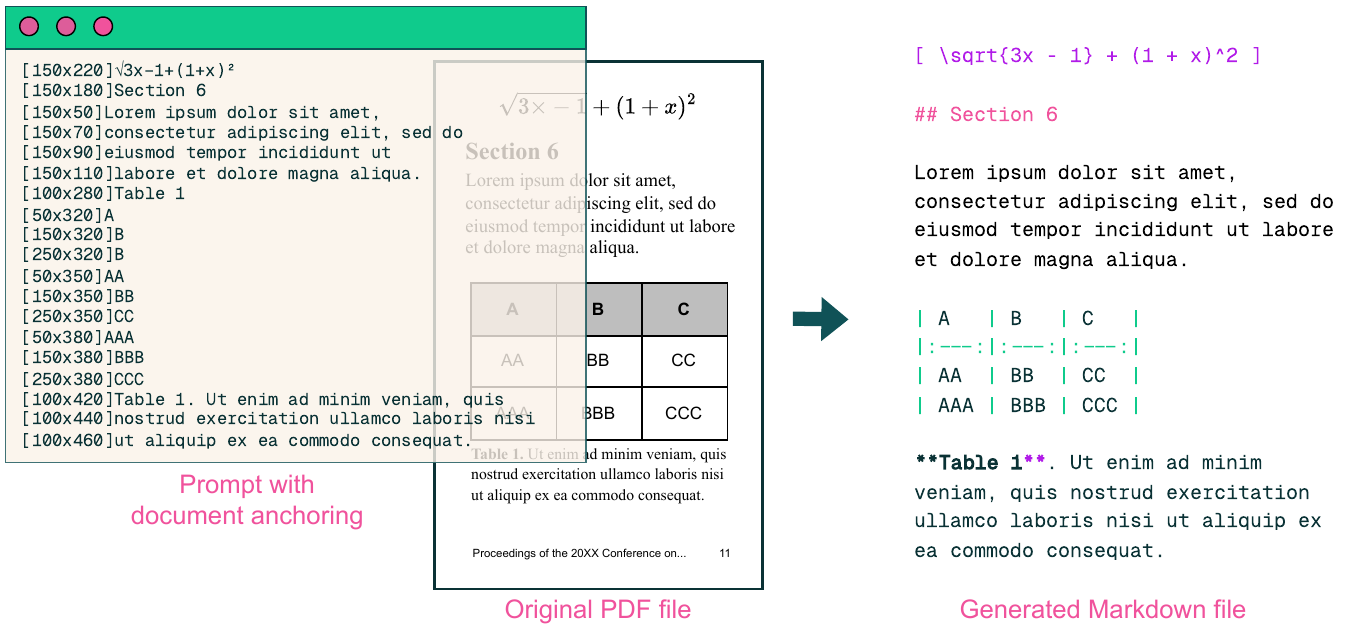}
    \label{fig:figure1}
\end{figure}

}

\abstract{
PDF documents have the potential to provide trillions of novel, high-quality tokens for training language models.
However, these documents come in a diversity of types with differing formats and visual layouts that pose a challenge when attempting to extract and faithfully represent the underlying content for language model use.
Traditional open source tools often produce lower quality extractions compared to vision language models (VLMs), but reliance on the best VLMs can be prohibitively costly (e.g., over \$6,240 USD per million PDF pages for GPT-4o) or infeasible if the PDFs cannot be sent to proprietary APIs.
We present \pipeline{}, an open-source toolkit for processing PDFs into clean, linearized plain text in natural reading order while preserving structured content like sections, tables, lists, equations, and more.
Our toolkit runs a fine-tuned 7B vision language model (VLM) trained on \train, a sample of 260,000 pages from over 100,000 crawled PDFs with diverse properties, including graphics, handwritten text and poor quality scans.
\pipeline{} is optimized for large-scale batch processing, able to scale flexibly to different hardware setups and can convert a million PDF pages for only \$176 USD.
To aid comparison with existing systems, we also introduce \bench, a curated set of 1,400 PDFs capturing many content types that remain challenging even for the best tools and VLMs, including formulas, tables, tiny fonts, old scans, and more.
We find \pipeline outperforms even top VLMs including GPT-4o, Gemini Flash 2 and Qwen-2.5-VL.
We openly release all components of \pipeline{}: our fine-tuned VLM model, training code and data, an efficient inference pipeline that supports vLLM and SGLang backends, and benchmark \bench.
}

\setmaintable{
\begin{table}[!h]
    \centering
    \begin{tabular}{l l l l l l}
        \github~{\sans{Code}} & \href{https://github.com/allenai/olmocr}{\texttt{allenai/olmocr}} & \huggingface~{\sans Weights \& Data} & \href{https://huggingface.co/collections/allenai/olmocr-67af8630b0062a25bf1b54a1}{\texttt{allenai/olmocr}} & \aitwo~{\sans{Demo}} & \href{https://olmocr.allenai.org/}{\texttt{olmocr.allenai.org}}
    \end{tabular}
\end{table}
}

\begin{document}

\maketitle

\section{Introduction}
\label{sec:intro}

Access to clean, coherent textual data is a crucial component in the life cycle of modern language models (LMs).
During model development, LMs require training on trillions of tokens derived from billions of documents~\citep{soldaini2024dolma,penedo2024fineweb,li2024datacomp}; errors from noisy or low fidelity content extraction and representation can result in training instabilities or even worse downstream performance~\citep{Penedo2023TheRD,li2024datacomp,OLMo2}.
During inference, LMs are often prompted with plain text representations of relevant document context to ground user prompts; for example, consider information extraction~\citep{Kim2021OCRFreeDU} or AI reading assistance~\citep{semantic-reader} over a user-provided document and cascading downstream errors due to low quality representation of the source document.

While the internet remains a valuable source of textual content for language models, large amounts of content are not readily available through web pages.
Electronic documents (\textit{e.g.}, PDF, PS, DjVu formats) and word processing files (\textit{e.g.}, DOC, ODT, RTF) are widely-used formats to store textual content.
However, these formats present a unique challenge: unlike modern web standards, they encode content to facilitate rendering on fixed-size physical pages, at the expense of preserving logical text structure.
For example, consider the PDF format, which originated as a means to specify how digital documents should be printed onto physical paper.
As seen in Figure~\ref{fig:pdf-chars}, PDFs store not units of text---headings, paragraphs, or other meaningful prose elements---but single characters alongside their spacing, placement, and any metadata used for visual rendering on a page.
As more and more documents became digital, users have relied this file format to create trillions of documents~\citep{pdfa2015};
yet, these documents remain difficult to leverage in LM pipelines because PDFs lack basic structure necessary for coherent prose, such as ground truth reading order.

\begin{figure}[t]  
    \centering
    \includegraphics[width=0.9\linewidth]{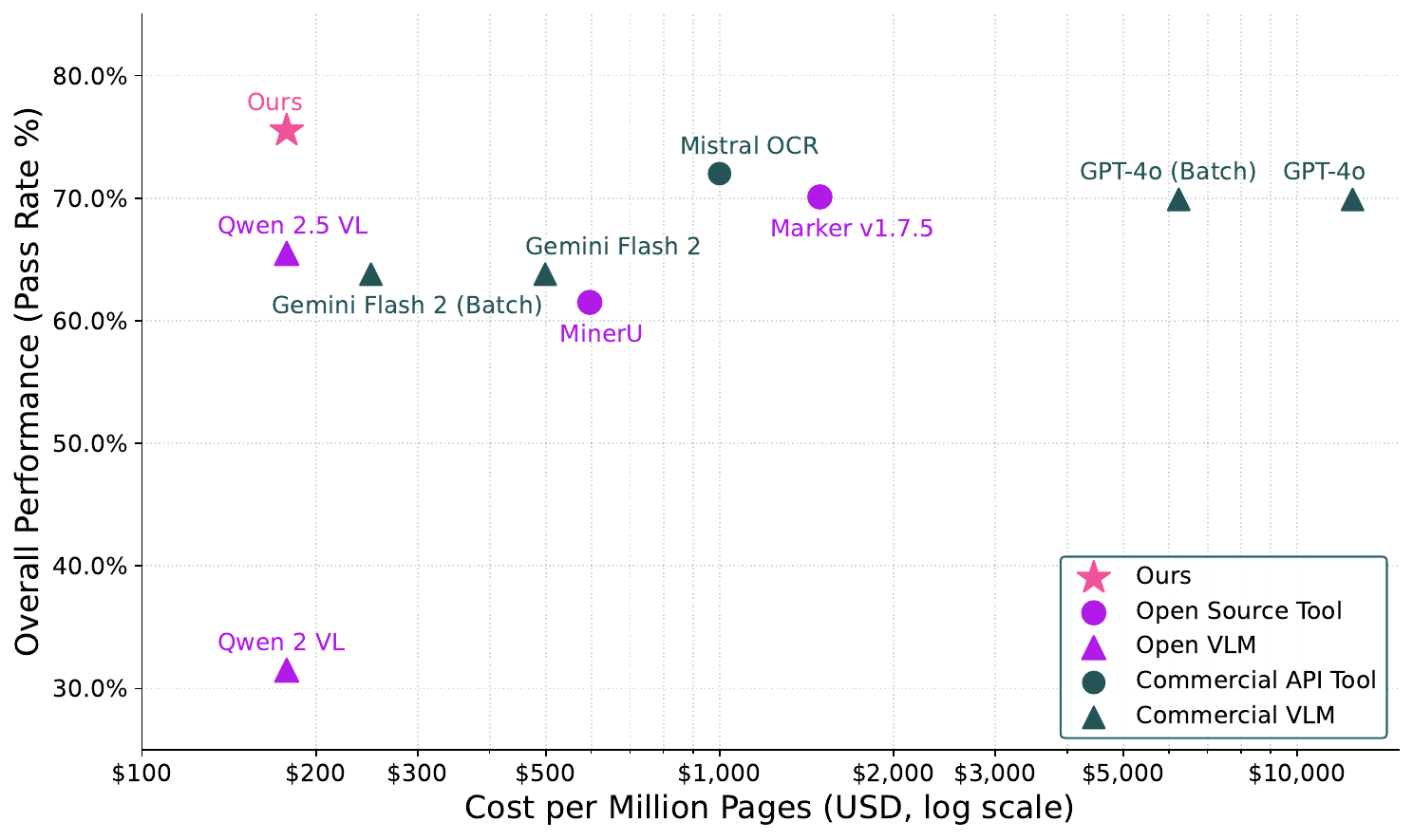}
    \caption{Performance-to-cost of \olmocr compared to a range of methods for PDF linearization and content extraction. Baselines include open and closed-source {\bf specialized tools} and {\bf general VLMs} prompted to perform this task.
    Performance is calculated on \bench, while Cost is calculated using commercial API pricing or the L40S GPU hourly rate (full details in Appendix~\ref{app:costs}). As \olmocr uses a fine-tuned Qwen 2 VL model (7B), they share the same inference cost; performance differences are a result of fine-tuning on our dataset \train.}
    \label{fig:pareto}
\end{figure}

\begin{figure}[!h]
    \centering
    \small
    \begin{tcolorbox}[
        colback=ai2offwhite,     
        colframe=ai2darkgreen,         
        coltitle=ai2offwhite,         
        fonttitle=\bfseries,      
        boxrule=1pt,              
        arc=3mm,                  
        boxsep=5pt                
    ]
    \begin{verbatim}
    Character: `o'
    Transform Matrix: (1.02, 0.0, 0, 1, 70.866, 709.481)
    Font: JURTWD+Manrope-Bold, Size: 24.78710000000001
    ------------------------------
    Character: `l'
    Transform Matrix: (1.02, 0.0, 0, 1, 86.490796356, 709.481)
    Font: JURTWD+Manrope-Bold, Size: 24.78710000000001
    ------------------------------
    Character: `m'
    Transform Matrix: (1.02, 0.0, 0, 1, 93.56999211600001, 709.481)
    Font: JURTWD+Manrope-Bold, Size: 24.78710000000001
    ------------------------------
    Character: `O'
    Transform Matrix: (1.02, 0.0, 0, 1, 116.299267074, 709.481)
    Font: JURTWD+Manrope-Bold, Size: 24.78710000000001
    ------------------------------
    Character: `C'
    Transform Matrix: (1.02, 0.0, 0, 1, 135.236115732, 709.481)
    Font: JURTWD+Manrope-Bold, Size: 24.78710000000001
    ------------------------------
    Character: `R'
    Transform Matrix: (1.02, 0.0, 0, 1, 153.894853128, 709.481)
    Font: JURTWD+Manrope-Bold, Size: 24.78710000000001\end{verbatim}
    \end{tcolorbox}
    \caption{Example of how PDFs represent textual content, such as this paper title, as individual glyphs with metadata.}
    \label{fig:pdf-chars}
\end{figure}

Faithful content extraction and representation of digitized print documents has long been of interest, with early research efforts in the 1950s, and first commercial optical character recognition (OCR) tools debuting in the late 1970s~\citep{Mori1992-qy}.
The release of Tesseract in 2006 represented a significant milestone, as the first high-quality, open-source OCR toolkit~\citep{Smith2013-pp}.
The current landscape of PDF extraction toolkits can be partitioned in pipeline-based systems and end-to-end models.
Pipeline-based systems (MinerU, \citealt{mineru}; Marker, \citealt{marker})  are comprised of multiple ML components (\textit{e.g.}, section segmentation, table parsing) chained together;
some, such as Grobid~\citep{grobid_01}, VILA~\citep{shen-etal-2022-vila}, and PaperMage~\citep{lopapermage}, are tailored to scientific papers.
On the other hand, end-to-end models parse a document with a single model.
For example, Nougat~\citep{nougat} and GOT Theory 2.0~\citep{gottheory} take images of PDF pages as input, and return plain text.
Notably, while pipeline-based systems have historically focused on simply faithful extraction, end-to-end-systems have also made strides to enable \emph{linearization} of this content---prescribing a flattening of this content to adhere to logical reading order---which can be quite challenging for layout-rich documents with many floating elements (e.g. multi-column documents with floating diagrams, headers, footnotes, and more).
Recently, rapid advances in the proprietary LMs have led to significant improvements in end-to-end text extraction capabilities~\citep{Bai2025-ub,geminiPDF}.
However, this capability comes at a steep price: for example, converting a million pages using GPT-4o can cost over \$6,200 USD.\footnote{With batch pricing, at \$1.25 USD (input) and \$5.00 USD (output) per million tokens in Feb 2025.}

\label{sec:intro:contributions}
We introduce {\bf\olmocr}, a general-purpose context extraction and linearization toolkit to convert PDFs or images of documents into clean plain text suitable for language model development.
Our contributions in this work are as follows:

\begin{itemize}[left=0pt,itemsep=0pt,topsep=0pt]
    \item \textbf{Data.} We create \href{https://huggingface.co/datasets/allenai/olmOCR-mix-0225}{\train}, a collection of 260,000 crawled PDF pages paired with their OCR output by GPT-4o, that we use to train our models.
    These documents represent a diverse set of publicly available PDFs, with a skew towards academic papers, public domain books, legal documents, brochures, and more.

    \item \textbf{Benchmark.} We develop \href{https://huggingface.co/datasets/allenai/olmOCR-bench}{\bench}, a comprehensive benchmark for evaluating document extraction tools. Unlike existing evaluation methods, \bench{} uses simple, natural binary rules, like software unit tests, that enable direct comparisons across different OCR systems without relying on fuzzy gold reference matching or LLM-as-judge for evaluation. The benchmark covers 1,400 PDF pages with over 7,000 unit-test cases spanning diverse document types.

    \item \textbf{Model and Code.} We fine-tune Qwen2-VL-7B-Instruct~\citep{qwen2vl} on \train{}, producing \href{https://huggingface.co/allenai/olmOCR-7B-0225-preview}{\model}.
    We package our VLM in the \olmocr Python toolkit, written to scale efficiently from one to hundreds of GPUs using SGLang~\citep{zheng2024sglang} and vLLM~\citep{kwon2023efficient} inference engines.
    \olmocr achieves state-of-the-art performance on our benchmark, even outperforming Qwen-2.5-VL-7B while remaining more cost-effective than existing alternatives, including commercial APIs; \olmocr can produce high-quality plain text at less than \$176 per million PDF pages.

    \item \textbf{Downstream Use.} We demonstrate real-world impact by applying \olmocr{} to process the 7.9M original PDFs in peS2o~\citep{pes2o_2023}, a widely-used corpus of linearized scientific articles used in language model pretraining. We show that training on the newly extracted data called \href{https://huggingface.co/datasets/allenai/olmOCR-pes2o-0225}{\pesto} can improve language model pretraining, observable even in downstream benchmark performance.
\end{itemize}

\section{Creating and Training on olmOCR mix}
\label{sec:model:dataset}
We face two challenges in data acquisition necessary for developing a VLM for our task: (1) acquiring a large, diverse set of PDFs and (2) obtaining their linearized plain text as supervision targets.

\subsection{Crawling PDFs}

\begin{table}[!h]
    \centering
    \small
    \begin{minipage}{0.45\textwidth}
        \begin{center}
        \begin{tabular}{lrr}
            \toprule
            \textbf{Source} & \begin{tabular}[c]{@{}r@{}}\textbf{Unique}\\\textbf{docs}\end{tabular} & \begin{tabular}[c]{@{}r@{}}\textbf{Total}\\\textbf{pages}\end{tabular} \\
            \midrule
            Web crawled PDFs & 96,929 & 240,940 \\
            Internet Archive books & 5,896 & 17,701 \\
            \midrule
            {\emph{Total}} & {\emph{102,825}} & {\emph{258,641}} \\
            \bottomrule
        \end{tabular}
        \caption{\train composition by source. Web crawled PDFs are sampled from a set of over 240 million documents crawled from public websites. Books in the Internet Archive set are in the public domain.}
        \label{tab:training_source}
        \end{center}
    \end{minipage}\hfill
    \begin{minipage}{0.5\textwidth}
        \centering
        \small
        \begin{tabular}{lr}
            \toprule
            \textbf{Document type} & \textbf{Fraction} \\
            \midrule
            Academic & 55.9\% \\
            Brochure & 11.2\% \\
            Legal & 10.2\% \\
            Books & 6.8\% \\
            Table & 5.6\% \\
            Diagram & 4.7\% \\
            Slideshow & 1.9\% \\
            Other & 3.7\% \\
            \bottomrule
        \end{tabular}
        \caption{\train PDFs breakdown by document type. Estimated by sampling 707 pages, classified using \texttt{gpt-4o-2024-11-20}. Prompt in Appendix~\ref{sec:classification_prompt}.}
        \label{tab:pdf_types}
    \end{minipage}
\end{table}

We randomly sample PDFs from an internal dataset of 240 million PDFs crawled from public internet sites, as well as PDFs of public domain books sourced from the Internet Archive.
While the web crawled set is often born-digital documents, PDFs from the Internet Archive consist of image scans.
We then perform a set of filters:
Using the Lingua package~\citep{lingua-py}, we identify and filter out non-English documents.
Further, we remove any document that failed to be parsed by \texttt{pypdf}, contains spam keywords, is a fillable form, or whose text is too short.\footnote{An implementation of these heuristics is available on GitHub:~\href{https://github.com/allenai/olmocr/blob/cc1f476b3e6018978ed1c6ca0b20b41ee2586604/olmocr/filter/filter.py\#L14-L112}{\path{/olmocr/filter/filter.py#L14-L112}}.}
We then sampled (up to) three pages uniformly at random from each PDF.
We summarize the data distribution in Tables~\ref{tab:training_source} and~\ref{tab:pdf_types}.


\subsection{Generating Linearized Plain Text}

Obtaining supervision targets for converting PDF to plain text presents a fundamental challenge.
First, human annotation is prohibitively expensive and can be error-prone.
Second, existing tools that extract content from PDF internals don't work on document images, but also don't provide reliable ground truth due to extraction errors from brittle heuristics.
In this work, we turned to data generation using GPT-4o to reliably convert PDF pages to linearized plain text.\footnote{In October 2024, we evaluated several leading VLMs for data generation. Gemini 1.5 was eliminated due to frequent \texttt{RECITATION} errors (though this was resolved by February 2025), GPT-4o mini produced excessive hallucinations, and Claude Sonnet 3.5 was cost-prohibitive. We selected \texttt{gpt-4o-2024-08-06} as it offered the optimal balance of accuracy, reliability, and cost-efficiency in batch mode.}

Yet, GPT-4o does not produce sufficiently high-fidelity plain text on its own; for high-density pages or complex layouts, we found it is prone to omitting content, rewriting or completing content in a manner unfaithful to the original, or captioning images when not instructed to do so.
To help guide GPT-4o generations, we experiment with augmenting the visual input (PDF page raster) with text blocks and position information extracted from the page.
As mentioned in \ref{sec:method}, we refer to this approach as \method{}.

We use the \texttt{pypdf}~\citep{pypdf} library to extract a representation of the page's structure from the PDF's internal data.
We note that this representation is \textit{highly noisy}: reading order is not preserved and main content is interwoven with boilerplate text and PDF rendering-related artifacts.
We sample blocks from this long extraction to add to the prompt until maximum input length is exceeded;
we prioritize text blocks and images which are located at the start and end of the document.

Finally, we instruct GPT-4o to respond with structured output to our requests.
We report the full JSON schema in Appendix~\ref{sec:json_schema}.
This forces the model to first extract page metadata, such as language, page orientation, and presence of tables, \emph{before} generating the text of the page in a natural reading order.
This format allows for more efficient processing of output;
further, we found it crucial to ensure that GPT-4o does not generate captions of images when no text is present on the page.
Overall, we find \method indeed improves the output quality of GPT-4o according to our benchmark (\S\ref{sec:bench}) reported in Table~\ref{tab:olmocr-bench-results}.

\subsection{Model Training}
\label{sec:model:training}

\paragraph{Fine-tuning}
While \method could be used to prompt any language model, its performance may depend on the model (Table~\ref{tab:olmocr-bench-results}), making it best suited as a data generation technique.
This leaves open the question whether a smaller, specialized VLM  can be as accurate as optimized prompting of a larger, general-purpose model.

Starting from a Qwen2-VL-7B-Instruct checkpoint, we fine-tune \model on \train.
Training is implemented using Hugging Face's \texttt{transformers} library~\citep{wolf-etal-2020-transformers}.
We use an effective batch size of 4, learning rate of 1e-6, AdamW optimizer, and a cosine annealing schedule for 10,000 steps (roughly 1.2 epochs).\footnote{
We manually tuned hyperparameters alongside other data curation decisions (e.g. \method prompt) throughout development.
To support this iterative cycle, we relied on manual side-by-side evaluation as shown in Appendix Figure~\ref{fig:eval-template}.
}
We use single node with 8 x NVIDIA H100 (80GB) GPUs.
A single training run took 16 node hours, with all training experiments totaling 365 node hours.

During fine-tuning, we slightly alter the \method prompt, removing some instructions and shrinking the image size so that PDF pages
are rendered to a maximum dimension of 1024 pixels on the longest edge.
The simplified text prompt is in Appendix~\ref{sec:fine_tune_prompt}.
The prompt is capped to 6,000 characters, so a typical prompt uses about 1,000 tokens to encode a page image, 1,800 tokens for the anchor text, for about 3,000 total input tokens.
Each training example was truncated to 8,192 tokens to cover cases when the prompt was unusually large.
Loss was masked so only the final response tokens participated in the loss calculation.

We keep the same structured JSON output that was present in the outputs of \train{}. More training evaluations are noted in Appendix~\ref{app:train-eval}.

\section{Building \bench}
\label{sec:bench}
We develop \bench to systematically evaluate PDF linearization and content extraction performance across diverse tools and models.
\bench operates by assessing a series of predefined pass-or-fail ``unit-tests''---\emph{Given an input whole PDF, does the plain text output satisfy a specific property or contain a specific element?}
Each test is designed to be simple, unambiguous, and deterministically machine-verifiable.
This avoids reliance on model-based evaluators which can be biased towards favoring their own generations~\citep{panickssery2024llmevaluatorsrecognizefavor}.
It also avoids
use of soft metric (e.g., edit distance, ROUGE) comparisons against reference text which might fail to reveal fine-grained yet semantically important content extraction errors, as is the case with incorrect math formulas (e.g., $x^i$ vs $x_i$).
\bench comprises 1,402 distinct PDF documents derived from diverse source repositories, covered by 7,010 unique test cases.
Some test patterns apply to any document type (e.g., presence, absence, reading order) while others are motivated by particular challenging yet important content extraction targets (e.g., tables, math formulas); see Table~\ref{tab:bench-sources} for a breakdown.

\subsection{Unit Test Categories}
We designed 5 distinct test categories, each designed to assess specific aspects of linearization and context extraction performance. We describe the test definitions and scoring methods below:

\begin{itemize}[left=0pt,itemsep=0pt,topsep=0pt]
\item \textbf{Text Presence}: Verifies that a text segment (typically spanning 1-3 sentences) is correctly identified within the plain text output. Soft/fuzzy matching is allowed, as well as specifying if the text must be in the first $N$ or last $N$ characters of the document. Case-sensitive by default.

\item \textbf{Text Absence}: Verifies that a text segment is successfully excluded from the plain text output. This category primarily targets peripheral content such as recurring headers, footers, and pagination markers. Soft/fuzzy matching is allowed, as well as specifying if the text must be in the first $N$ or last $N$ characters of the document. \textit{Not} case-sensitive by default.

\item \textbf{Natural Reading Order}: Verifies the order between two text segments. For instance, on a PDF with multiple news articles on one page, we can test for whether the first sentence of the first article appears after the heading of that article; yet such tests can be designed to not penalize for the order of the articles themselves.
Soft matching is allowed, case-sensitive by default.

\item \textbf{Table Accuracy}: Checks that the plain text output contains a table with a cell with a given value, and that its neighboring cells have certain properties. For instance, one can validate this page has a table with a cell containing ``4.5\%'' and above that is a cell containing ``2.4\%''. Both Markdown and HTML based tables are supported, though many cases depend on \texttt{rowspan} and \texttt{colspan} information being preserved, which is possible only in HTML based tables.

\item \textbf{Math Formula Accuracy}: Checks that the plain text output contains a given math equation. We render a reference \LaTeX{} equation using KaTeX in a headless browser and extract all rendered symbols and their (visual) bounding boxes.
Then we check if a matching collection of symbols, with the same relative orientations, exists anywhere in the final OCR document. For instance, if searching for $f(x) = \int_{-3}^{3} x^2 dx$ on a page, we look for an equation where $\int$ appears to the left of a $x$, $x$ appears to the left of $dx$, $3$ appears above $-3$, and so on.
This is similar to the method described by~\citet{cdm}, but ours is simpler due to the test being Pass/Fail only.

\item \textbf{Baseline}: Each PDF document by default also receives a baseline or default test case, which checks that some plain text output containing alphanumeric characters was actually produced for that page, that such output does not have a string of repeating N grams at the end (longer than 30), and that
the output does not contain any characters from the Chinese, Japanese, or Emoji Unicode charsets.\footnote{This is to test for common failure cases of models, such as accidentally switching languages and generating repeated outputs. The handful of pages which do legitimately contain such charsets are manually flagged and excluded from such test conditions.}

\end{itemize}

In all cases where text is compared, we perform basic string normalization, such as converting \textit{<br>}s to newlines, normalizing all whitespace to single ASCII spaces, removing Markdown bold/italics, normalizing quotes/hyphens to ASCII, and converting all Unicode to NFC format.

\begin{table}[!t]
\begin{center}
\begin{tabular}{lrrrrrr}
\toprule
& Presence & Absence & Read Order & Table & Formula & Total tests \\
\midrule
arXiv Math (AM) & - & - & - & - & 2,927 & 2,927 \\
Old Scans Math (OSM) & - & - & - & - & 458 & 458 \\
Tables (TA) & - & - & - & 1,020 & - & 1,020 \\
Old Scans (OS) & 279 & 70 & 177 & - & - & 526 \\
Headers Footers (HF) & - & 753 & - & - & - & 753 \\
Multi Column (MC) & - & - & 884 & - & - & 884 \\
Long Tiny Text (LTT) & 442 & - & - & - & - & 442 \\
\midrule
Total Tests & 721 & 823 & 1,061 & 1,020 & 3,385 & 7,010 \\
\bottomrule
\end{tabular}
\caption{Counts of unit test types in \bench.}
\label{tab:bench-sources}
\end{center}
\end{table}

\subsection{Sourcing Documents and Creating Tests}

We define 7 distinct document types that we found \olmocr (or its earlier iterations) often struggled to process and defined custom acquisition strategies for each (described below).
We removed documents that both contained PII and were not meant for public dissemination; prompt in Appendix~\ref{sec:pii}.
We also decontaminate against documents that appear in \train{} via URL level deduplication.
To scale creation of test cases over these documents, we combined manual design and review with prompting GPT-4o; further details and prompts are in Appendix~\ref{appendix:bench01}.
Visualize sample documents in Appendix~\ref{sec:samples}.

\begin{itemize}[left=0pt,itemsep=0pt,topsep=0pt]

\item \textbf{arXiv Math (AR)}
We downloaded a recent set of papers from the math subset of arXiv, selecting manuscripts with a single TeX source file and corresponding rendered PDF.
To select a candidate \LaTeX{} expression from a page to use in a test, we (1) ran \pipeline{} to identify candidate pages with TeX, (2) match pages back to original TeX source, and (3) validate matched TeX rendering compatibility with KaTeX.

We manually verify the final set of test cases to exclude instances where custom macros produce renderings that deviate from standard \LaTeX{} and to split multi-part equations into smaller test cases.

\item \textbf{Old Scans Math (OSM)}
We crawl old, public domain math textbooks from the Internet Archive\footnote{https://archive.org}, extracting random pages from these documents. We similarly use \pipeline{} to find candidate pages with formulas, but this time manually annotate each formula on the page to use as test cases.

\item \textbf{Tables (TA)}
We sampled more documents from the same internal crawled PDF repository used to create \train and filtered to those which had tables using a simple prompt with Gemini-Flash-2.0.
On pages with tables, we prompted Gemini-Flash-2.0 for the relationships between randomly chosen cells.
We manually reviewed those tests for accuracy.

\item \textbf{Old Scans (OS)}
We sampled historical letters and typewritten documents with existing human transcriptions from the Library of Congress\footnote{https://crowd.loc.gov} digital archives.
We then wrote a small script to generate Natural Reading Order cases consisting of sentences that were naturally before or after one another in the original human transcriptions. We manually added test cases to cover some headers/footers which should have been excluded from any OCR version of these documents.
All of the test cases then underwent a second pass of human review for accuracy.

\item \textbf{Headers Footers (HF)}
We sampled documents from the same internally crawled PDF repository as \train. We used DocLayout-YOLO~\citep{zhao2024doclayout} to identify page regions labeled as headers or footers using the \texttt{abandon} category. To extract the text from these header/footer regions, we visually mask out the rest of the document and prompt Gemini-Flash-2.0 for the content. These extracted snippets are added as test cases that should be absent in linearized output.
We manually reviewed to remove mistakenly filtered text and to set conditions such as limiting the search area to the first N or last N characters. For example, if a page number ``5'' appears at the bottom of a page, we test to ensure that output plain text does not contain ``5'' in the last 20 characters, but still allow for a ``5'' that may appear earlier in the text.

\item \textbf{Multi Column (MC)}
We visually sample documents from our internal crawled PDF repository to find documents with multi-column layouts and multiple articles on one page. We use Claude-Sonnet-3.7 to render those pages to HTML, and from that HTML, we extract text segments before/after one another.
We manually review each entry for accuracy. We purposely select simple text blocks from coherent regions of the document, and avoid including any math formulas, superscripts, or subscripts in these tests.

\item \textbf{Long Tiny Text (LTT)}
We crawled documents from the Internet Archive containing a large amount of dense, small print on a single page. Such documents include pages from a dictionary or pages of references from academic papers.
We then generate test cases using Gemini-Flash-2.0 and verify them manually.
\end{itemize}


\subsection{Scoring}
We run each of the PDF pages across each of our tools and methods to produce a markdown or plain text document.
As all tests are Pass/Fail, we simply report percentage of tests passed, macro-averaged by document type.
We evaluate each of the tests to get a percentage correct score for each test source (plus the default baseline tests).
The final score for each tool is the average of the percentage across each test source. This captures the difficulty we faced at times of finding and validating enough cases from each source, but we roughly feel that each source represents an important capability for an OCR system to have.

\[
\text{Overall score} = \frac{1}{N} \sum_{s \in \text{Document sources}} \text{Score}(s)
\]




\section{Evaluating \olmocr}
\label{sec:results}

First, we evaluate \olmocr{} on \bench{} against a range of linearization tools and VLMs (Section~\S\ref{sec:result:bench_results}).
We then quantify the usefulness \olmocr for language modeling by continued pretraining on an OLMo 2 checkpoint~\citep{OLMo2} on content extracted and linearized with our toolkit (Section~\S\ref{sec:results:annealing}).

Additional evaluations, studying how faithful \olmocr is to its teacher model (Section~\S\ref{sec:result:alignment}), and pairwise ELO comparison (Section~\S\ref{sec:elorating}) are available in the appendix.

\subsection{\bench Results}
\label{sec:result:bench_results}
From Table~\ref{tab:olmocr-bench-results}, we see that \olmocr significantly outperforms both the best commercial dedicated OCR tool (Mistral) as well as both GPT-4o, its teacher model, and Qwen 2.5 VL, which is an update to Qwen 2 VL, which was the base model for \model.
We note that we developed \bench{} \emph{after} training \model to prevent unfairly iterating on the benchmark before comparing with other methods.
Qualitatively, \olmocr produces significantly cleaner plain text than specialized open-source tools (visualized in Appendix~\ref{appendix:olmocr_output}).

\begin{table}[ht]
\centering
\small
\begin{tabular}{lcccccccc@{\hspace{2em}}c}
\toprule
\textbf{Model} & {AR} & {OSM} & {TA} & {OS} & {HF} & {MC} & {LTT} & {Base} & {Overall} \\
\midrule
GOT OCR                    & 52.7 & 52.0 & 0.2  & 22.1 & 93.6  & 42.0  & 29.9  & 94.0  & 48.3 ± 1.1 \\
Marker v1.7.5              & 76.0 & 57.9 & 57.6 & 27.8 & 84.9  & 72.9  & 84.6  & 99.1 & 70.1 ± 1.1 \\
MinerU v1.3.10             & 75.4 & 47.4 & 60.9 & 17.3 & \textbf{96.6} & 59.0  & 39.1  & 96.6  & 61.5 ± 1.1 \\
Mistral OCR API            & \textbf{77.2} & 67.5 & 60.6 & 29.3 & 93.6  & 71.3  & 77.1  & \textbf{99.4}  & 72.0 ± 1.1 \\
\midrule
GPT-4o (No Anchor)         & 51.5 & \textbf{75.5} & 69.1 & 40.9 & 94.2  & 68.9  & 54.1  & 96.7  & 68.9 ± 1.1 \\
GPT-4o (Anchored)          & 53.5 & 74.5 & 70.0 & 40.7 & 93.8  & 69.3  & 60.6  & 96.8  & 69.9 ± 1.1 \\
Gemini Flash 2 (No Anchor) & 32.1 & 56.3 & 61.4 & 27.8 & 48.0  & 58.7  & \textbf{84.4} & 94.0  & 57.8 ± 1.1 \\
Gemini Flash 2 (Anchored)  & 54.5 & 56.1 & \textbf{72.1} & 34.2 & 64.7  & 61.5  & 71.5  & 95.6  & 63.8 ± 1.2 \\
Qwen 2 VL (No Anchor)      & 19.7 & 31.7 & 24.2 & 17.1 & 88.9  & 8.3   & 6.8   & 55.5  & 31.5 ± 0.9 \\
Qwen 2.5 VL (No Anchor)    & 63.1 & 65.7 & 67.3 & 38.6 & 73.6  & 68.3  & 49.1  & 98.3  & 65.5 ± 1.2 \\
Ours (v0.1.75 No Anchor)           & 71.5 & 71.4 & 71.4 & \textbf{42.8} & 94.1  & 77.7  & 71.0  & 97.8  & 74.7 ± 1.1 \\
Ours (v0.1.75 Anchored)            & 74.9 & 71.2 & 71.0 & 42.2 & 94.5  & \textbf{78.3} & 73.3  & 98.3  & \textbf{75.5 ± 1.0} \\
\bottomrule
\end{tabular}
\caption{Evaluation results on \bench grouped by document types. Best unit test pass rate in each column is bold. 95\% CI calculated by bootstrapping with 10k samples.}
\label{tab:olmocr-bench-results}
\end{table}

\subsection{Downstream Evaluation}
\label{sec:results:annealing}

To assess the impact of improved PDF linearization, we experiment using an intermediate checkpoint of \texttt{OLMo-2-1124-7B} and continued pretraining using content extracted from a fixed collection of PDFs but with different linearization tools.
This ablation procedure has been used to assess data quality in~\citep{blakeney2024doesdatasparkjoy,dubey2024llama,OLMo2}.

For our baseline, we use PDF extracted tokens from \texttt{peS2o}~\citep{pes2o_2023}, 58B tokens from academic papers derived using Grobid~\citep{grobid_01} from the S2ORC~\citep{lo-etal-2020-s2orc} paper collection and further cleaned with heuristics for language modeling.
To represent \pipeline{}, we identify the same documents used in peS2o, acquire their source PDFs from the upstream S2ORC pipeline, and reprocess them using \pipeline{}.
For these two versions of peS2o, we train the 7B checkpoint for another 50B tokens. As shown in Table \ref{tab:downstream}, replacing the original peS2o tokens extracted via \texttt{Grobid + rules} with those processed used \pipeline{} results in a +1.3 percentage point average improvement on widely-reported LM benchmark tasks, including MMLU~\citep{hendrycks2021measuringmassivemultitasklanguage}, ARC\textsubscript{C}, DROP~\citep{dua2019dropreadingcomprehensionbenchmark}, HellaSwag~\citep{zellers2019hellaswagmachinereallyfinish}, NaturalQuestions~\citep{kwiatkowski-etal-2019-natural}, WinoGrande~\citep{sakaguchi2019winograndeadversarialwinogradschema}.

\begin{table}[!h]
\begin{center}
\begin{tabular}{lcccccccc}
\toprule
\textbf{peS2o version} & \textbf{Average} & \textbf{MMLU} & \textbf{$\textbf{ARC}_\text{\scriptsize\bf{C}}$} & \textbf{DROP} & \textbf{HSwag} & \textbf{NQ} & \textbf{WinoG} \\
\midrule
$\text{Grobid}+\text{rules}$~\href{https://huggingface.co/datasets/allenai/peS2o}{\citep{pes2o_2023}}
& 53.9 & {\bf{61.1}} & 75.0 & 42.3 & 57.4 & {\bf{29.4}} & {\bf{58.3}} \\
\pesto & {\bf{55.2}} & {\bf{61.1}} & {\bf{76.4}} & {\bf{43.7}} & {\bf{62.6}} & 29.1 & 58.0 \\
\bottomrule
\end{tabular}
\caption{Comparison on \textsc{OLMo} 2~\citep{OLMo2} downstream evaluation tasks of \texttt{OLMo-2-7B-1124} on 50B of original peS2o tokens vs 50B tokens from the same source PDFs but processed with \pipeline{}.}
\label{tab:downstream}
\end{center}
\end{table}

\subsection{Cost Evaluation}
\label{sec:results:cost}

Finally, when considering real-world use, cost efficiency is just as important as performance.
We present a summary of inference costs in Table \ref{tab:inference_cost}.
To contextualize the value of \olmocr, at 1,000 tokens per page, to process all of peS2o PDFs can already cost \$10.3M in H100 usage.
In comparison, Mistral OCR is a commercial API tool specializing in this task, yet is over five times more expensive, making it even more prohibitive to use for language modeling.
See Appendix~\ref{app:costs} for details on pricing and cost calculations.

\begin{table}[!h]
    \centering
    \begin{tabular}{ccccc}
        \toprule
        \textbf{Model} & \textbf{Hardware} & \textbf{Tokens/sec} & \textbf{Pages/USD} & \textbf{Cost per million pages}\\
        \midrule

        \multirow{2}{*}{GPT-4o} & API & - & 80 & \$12,480   \\
         & Batch & - & 160 & \$6,240 \\



        Marker v1.7.5 (Force OCR) & H100 & 332 & 674 & \$1484 \\

        Mistral OCR & API & - & 1,000 & \$1,000 \\

        MinerU & L40S & 238 & 1,678 & \$596 \\

        \multirow{2}{*}{Gemini Flash 2} & API & - & 2,004 & \$499  \\
         & Batch & - & 4,008 & \$249 \\

        \midrule

        \multirow{2}{*}{\pipeline}& L40S & 906 & {\bf{5,697}} & {\bf{\$176}} \\
         & H100 & 3,050 & {\bf{5,632}} & {\bf{\$178}}\\
        \bottomrule
    \end{tabular}
    \caption{Inference cost comparison against other OCR methods. NVIDIA L40S estimated at \$0.79 per hour, H100 80GB estimated at \$2.69 per hour.
    We measured a 12\% retry rate for \olmocr. Full cost breakdown in Appendix~\ref{app:costs}.
    }
    \label{tab:inference_cost}
\end{table}

\section{Related Work}
\label{sec:related-work}

\paragraph{Tools and Models for Linearizing PDFs to Plain Text.} Many tools have existed for this task, some are parsers of born-digital PDF internals while others are OCR tools on top of image rasters of PDF pages.
As machine learning matured, more people started developing models that automate this PDF parsing; examples include LayoutLM~\citep{Xu_2020_layputlm} and VILA~\citep{lin2024vilapretrainingvisuallanguage}.
Tools have been developed around use of these models, including PaperMage~\citep{lo2023papermage}, or have updated to include their own custom trained models, like Grobid~\citep{grobid_01}.
Commercial API providers began integrating VLMs with document processing capabilities: OpenAI introduced GPT-4 Vision~\citep{openai2024gpt4technicalreport} in September 2023, Google launched Gemini~\citep{geminiteam2025geminifamilyhighlycapable} in December 2023 with significant enhancements throughout 2024 as multimodal models became more powerful and accessible. Despite these developments, there remained a notable absence that specifically train a VLM for this task and package this capability into comprehensive, production-ready software libraries. Our work addresses this gap, developing alongside concurrent efforts such as Mistral~\citep{mistral_ai_2025}, Qwen VL~\citep{bai2023qwenvlversatilevisionlanguagemodel} which we systematically evaluate against.

\paragraph{Benchmarking VLMs on Linearization.} Several benchmarks have been developed for evaluating document understanding and content extraction. Established datasets such as FUNSD~\citep{jaume2019} focus on form understanding with typewritten content, SROIE~\citep{huang2019icdar2019} concentrates on information extraction from scanned receipts, and RVL-CDIP~\citep{harley2015icdar} contains scanned documents. However, these datasets exhibit significant limitations: they are predominantly domain specific, targeting narrow document categories with constraint formatting variations, while our approach leverages a diverse corpus spanning multiple domains and document types.
Additionally, traditional benchmarks often focus on isolated extraction tasks (e.g., exclusively evaluating tables with PubTabNet~\citep{zhong2020image}, or mathematical formulas with specialized detection frameworks~\citep{zhong20211stplacesolutionicdar}), whereas our benchmark evaluates performance across a comprehensive spectrum of extraction challenges. Further, their evaluation method is brittle, typically relying on exact string matching against predefined gold-standard tokens (which makes it difficult to compare methods that produce different tokenizations). In contrast, our unit-test-style evaluation framework enables equitable assessment across diverse implementations regardless of their underlying tokenization mechanisms, providing a more generalizable evaluation paradigm for document understanding systems.

\paragraph{Linearization for Language Modeling.} Significant work has been done on how to curate data for language modeling, with significant efforts on topics like data filtering~\citep{soldaini2024dolma,penedo2024fineweb,li2024datacomp} and source mixing~\citep{wettig2025organize,liu2024regmix}. However, relatively little attention has been directed toward understanding the impact of linearization processes on downstream model training. DCLM~\citep{li2024datacomp} and RefinedWeb~\citep{Penedo2023TheRD} touched on some aspects of this challenge, utilizing tools like Resiliparse~\citep{bevendorff:2018} and Trafilatura~\citep{barbaresi-2021-trafilatura}, but their approaches were restricted to web based textual content. OpenWebMath~\citep{paster2023openwebmath} showed accurate content extraction is important for specialized domains, like mathematical formulas.
For PDF-based content specifically, similar research contributions remain limited, which motivates this work.

\section{Conclusion}

We introduce \pipeline{}, an open-source toolkit for converting PDF documents into clean plain text.
Our approach combines \method{}, a novel prompting technique that leverages available metadata in born-digital PDFs, with a fine-tuned 7B parameter vision language model to achieve results competitive with closed commercial solutions at a fraction of the cost.
We openly release our training set \train to enable others to further develop their own VLMs.

To rigorously evaluate the system, we developed \bench, a benchmark of 7,010 test instances across 1,403 PDFs. It includes Pass/Fail unit tests for text presence, reading order, tables, formulas, and baseline functionality. The documents span categories from scientific papers to historical manuscripts, enabling robust assessment across diverse linearization and context extraction challenges.

Our released efficient inference pipeline contains everything needed to
start converting anything from single documents to million-page archives of PDFs.
We hope \pipeline{}'s ability to efficiently process millions of documents will help unlock new sources of training data for language models, particularly from high-quality PDF documents that are currently underrepresented in existing datasets that rely heavily solely on crawled web pages.

\section*{Acknowledgments}
\label{sec:acks}

This work would not be possible without the support of our colleagues at Ai2.  We thank Byron Bischoff, Aaron Sarnat, Huy Tran, Sam Skjonsberg, Eric Marsh, and Chris Newell for help setting up the live demo; Taira Anderson, Sruthi Sreeram for program management support; Will Smith and Crystal Nam for legal guidance; Michael Schmitz, Caitlin Wittlif and Carissa Schoenick for various indirect support.
We also thank Benjamin Charles Germain Lee for helpful feedback and suggestions on evaluation and potential use cases.
We are grateful for the extensive feedback provided by Hynek Kydl\'{i}\v{c}ek on the inference toolkit.

\bibliographystyle{ai2style/plainnat}
\bibliography{references}

\clearpage
\appendix

\section{Methodology}
\label{sec:method}

\paragraph{Approach}
Many end-to-end OCR models, such as GOT Theory 2.0~\citep{gottheory} and Nougat~\citep{nougat}, exclusively rely on rasterized pages to convert documents to plain text;
that is, they process images of the document pages as input to autoregressively decode text tokens.
This approach, while offering great compatibility with image-only digitization pipelines, misses the fact that most PDFs are born-digital documents, thus already contain either digitized text or other metadata that would help in correctly linearizing the content.

\begin{figure}[h]  
    \centering
    \includegraphics[width=0.8\linewidth]{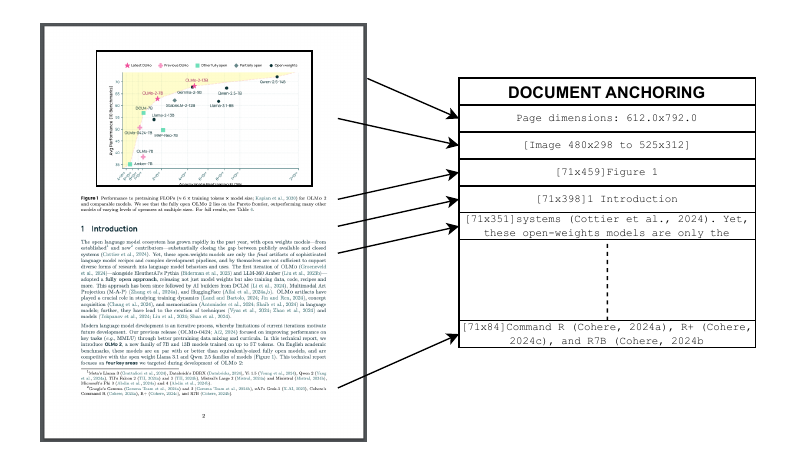}
    \caption{Example of how \method{} works for a typical page. Relevant image locations and text blocks get extracted, concatenated, and inserted into the model prompt.
    When prompting a VLM for a plain text version of the document, the anchored text is used in conjunction with the rasterized image of a page.}
    \label{fig:anchoring-example}
\end{figure}

In contrast, the \olmocr pipeline leverages document text and metadata.
We call this approach {\bf\method}.
Figure~\ref{fig:anchoring-example} provides an overview of our method;
\method extracts coordinates of salient elements in each page (\textit{e.g.}, text blocks and images) and injects them alongside raw text extracted from the PDF binary file.
Crucially, the anchored text is provide as input to any VLM \textit{alongside} a rasterized image of the page.

Our approach increases the quality of our content extraction.
We apply \method{} when prompting \GptFourO to collect silver training samples,
when fine-tuning \model,
and when performing inference with the \olmocr toolkit.

\paragraph{Implementation}
\method{} processes PDF document pages via the \texttt{pypdf}~\citep{pypdf} library to extract a representation of the page's structure from the underlying PDF.
All of the text blocks and images in the page are extracted, including position information.
Starting with the most relevant text blocks and images\footnote{We prioritize text blocks and images which are located at the start and end of the document.}, these are sampled and added to the prompt of the VLM, up to a defined maximum character limit\footnote{We use a character limit for convenience and speed, but during training or inference, if a page's prompt exceeds the model's token limit, we just regenerate it with exponentially decreasing character limits until it is suitable.}. This extra information is then available to the model when processing the document.

Overall, we find that using prompts constructed using \method results in significantly fewer hallucinations.
Prompting with just the page image was prone to models completing unfinished sentences, or to invent larger texts when the image data was ambiguous.
Finally, while \method{} helps with quality on born-digital documents, our pipeline maintains high performance on documents that do not have any digital metadata encoded in them.
In these cases, the model will not have the benefit of seeing the internal structure of the PDF document,
instead relying on just the rasterized image of a page to process the underlying document.

\section{Cost Estimates of PDF Extraction Systems}
\label{app:costs}
To estimate prices, we use rates provided by RunPod\footnote{\url{https://www.runpod.io}} as of February 2025. It prices a single on-demand NVIDIA L40S GPU at \$0.79 USD per hour, and NVIDIA H100 80GB SXM at \$2.69 USD per hour. Using these rates, costs (in USD) were computed as follows:

\begin{itemize}[left=0pt,itemsep=0pt,topsep=0pt]
    \item \textbf{GPT-4o}: We evaluated GPT-4o in February 2025. We tested  1288 pages, which resulted in 3,093,315 input tokens at 833,599 output tokens. Priced at \$2.50 per million input tokens and \$10.00 per million output tokens, it resulted in a total of \$16.07. Batch processing is priced at half of the cost, \$8.03.
    \item \textbf{Mistral OCR}: As of May 2025, Mistral prices their OCR service at \$1 per 1,000 pages, regardless of number of generated tokens.
    \item \textbf{MinerU}: We run the toolkit (version 1.3.10) on a single NVIDIA L40S GPU. It processed 1,288 pages in 58 minutes 22 seconds, costing \$0.767.
    \item \textbf{Marker}: We run marker v1.7.5 using the marker command line with the \texttt{force\_ocr} flag on 10,000 pages selected randomly from \train{}. This took 5 hours, 31 minutes on an H100 node with 1 GPU, resulting in a price of \$14.84 for 10,000 pages. \footnote{The force\_ocr option is more expensive, but results in better performance. The Marker authors are working on improving performance on large GPUs with multiple workers which should lower this cost.}
    \item \textbf{Gemini Flash 2.0}: As of February 2025, it is priced \$0.10 per 1 million input tokens, and \$0.40 per 1 million output tokens. In our testing over the same 1,288 pages used to evaluate GPT-4o, it cost in \$0.643.
    \item \textbf{\olmocr}: We tested the launch version of \olmocr on both L40S and H100 GPUs. On L40s, it processed 1,288 test pages in 17 minutes, 10 seconds. The effective throughput of the model was 906 output tokens per second, plus a 12\% reties rate. Overall, we estimate its costs at \$0.226. On H100, \olmocr generates 3,050 output tokens per second, resulting in a runtime of 5 minutes 7 seconds, for a cost of \$0.229.


\end{itemize}

\section{Evaluation of Trained Models}
\label{app:train-eval}

\begin{figure}[h]
    \centering
    \begin{minipage}{0.48\textwidth}
        \centering
        \includegraphics[width=\linewidth]{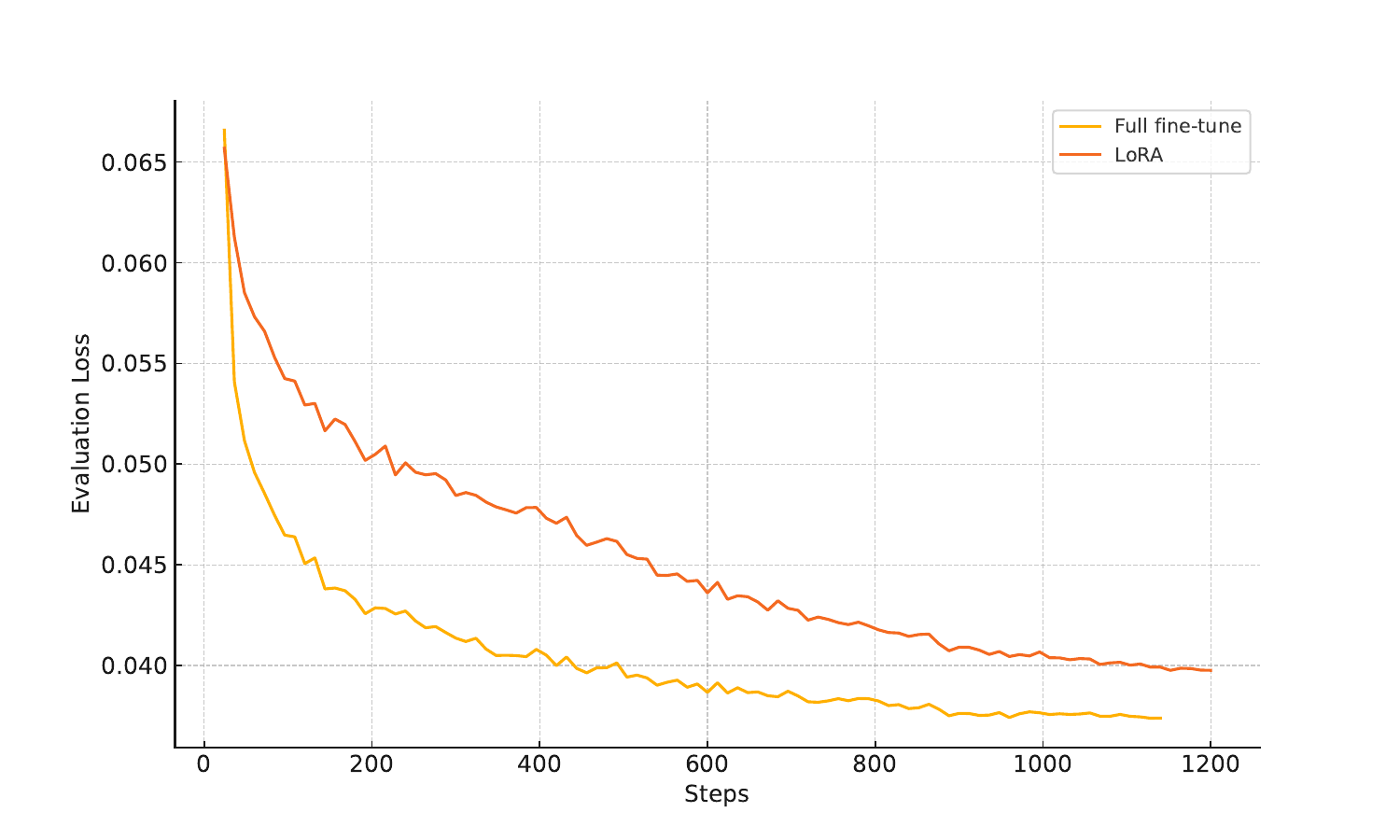}
        \caption{Validation Loss - Web PDFs}
        \label{fig:s2pdf-loss}
    \end{minipage}
    \hfill
    \begin{minipage}{0.48\textwidth}
        \centering
        \includegraphics[width=\linewidth]{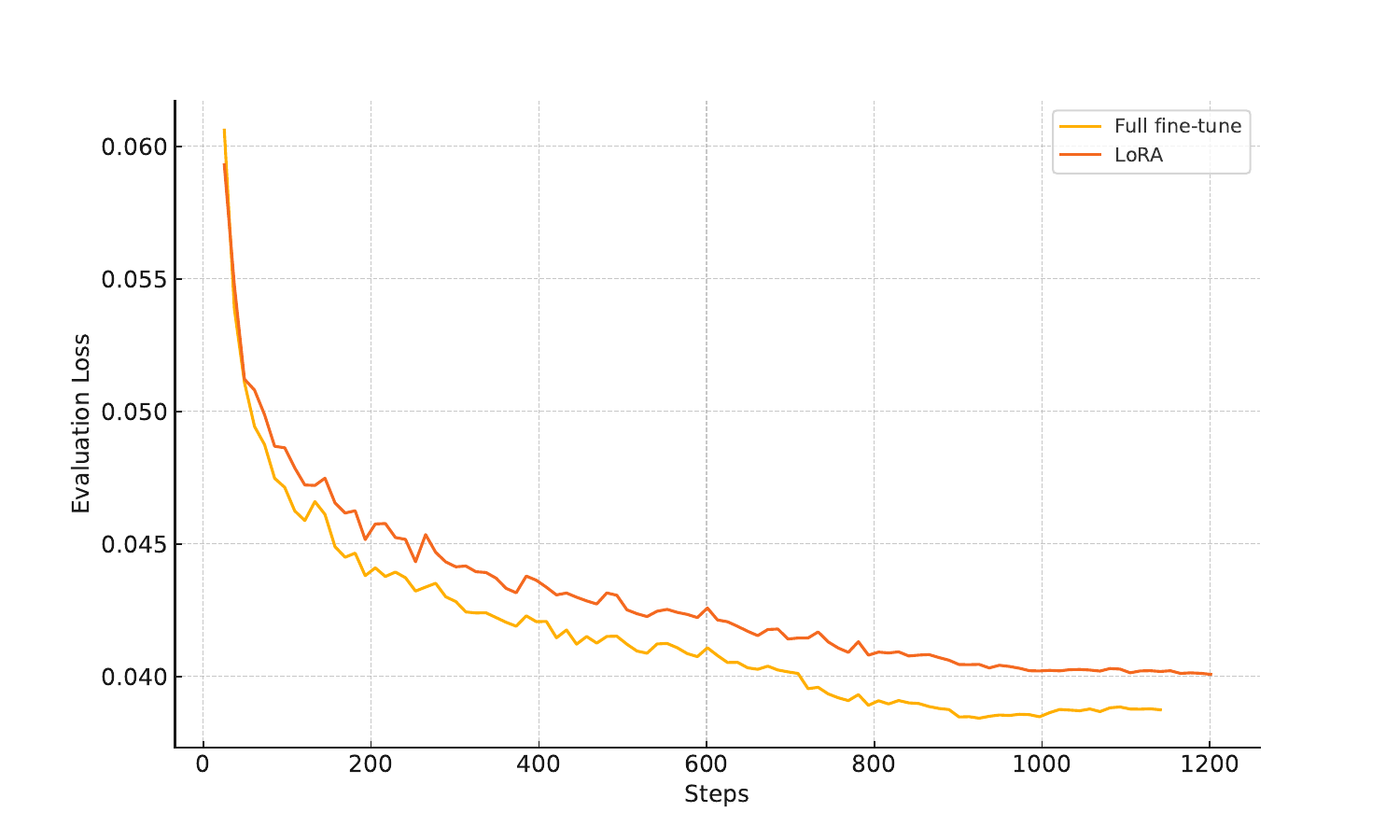}
        \caption{Validation Loss - Internet Archive Books}
        \label{fig:iabooks-loss}
    \end{minipage}
\end{figure}

We track validation loss during training of \model{} against a development subset of \train{} during fine-tuning;
Figure~\ref{fig:s2pdf-loss} and Figure~\ref{fig:iabooks-loss}, show the loss curves for both the web PDFs and the Internet Archive books subsets.
LoRA resulted in higher loss values compared to full fine-tuning, which we use for the final model.

\begin{figure}[!h]
    \centering
    \includegraphics[width=0.8\linewidth]{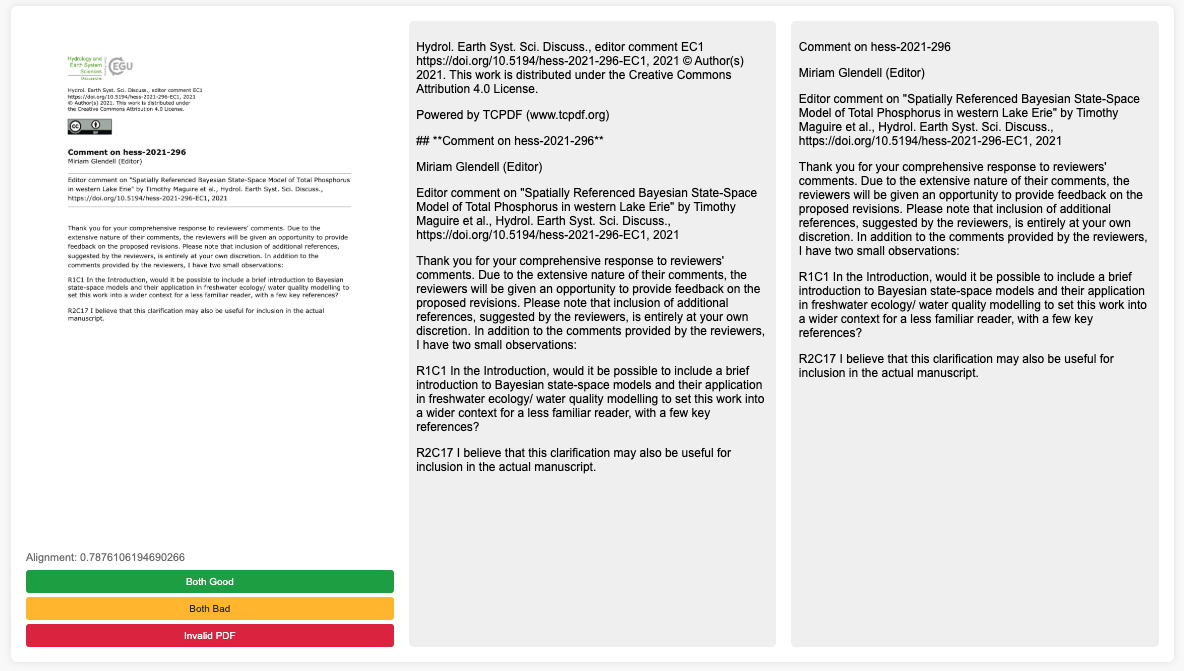}
    \caption{Example of side-by-side evaluation tool used during development.
    The software used to create these comparisons is released as open-source software as part of \pipeline{}.}
    \label{fig:eval-template}
\end{figure}

To set hyperparameters and make other decisions during development, we relied on manual side-by-side evaluation as shown in Figure~\ref{fig:eval-template}. A random selection of 20 to 50 documents
were processed using two different methods, and were displayed side by side along with the render of the document page. We also open source our evaluation tool to support qualitative inspection of this visually-rich data.

\subsection{Alignment with Teacher Model}
\label{sec:result:alignment}



To compare the output of \model{} to the GPT-4o silver data in \train{}, we build a document similarity
metric which splits a document into words, uses Hirschberg's algorithm to align those words,  and counts what proportion match.

We report alignment scores in Table~\ref{tab:eval_alignment}.
Overall, we find that \model{} has good alignment, 0.875 on average, with its teacher model.
To calibrate this result, we also report GPT-4o self-alignment score of 0.954, which is simply from calling the model again; imperfect alignment here is due to resampling differences.
In fact, we find that our model actually better mimics the content extraction and linearization of GPT-4o than its smaller counterpart GPT-4o mini.

When partitioning scores in low, medium, and high alignment buckets (Table~\ref{tab:eval_matching}), we find that most documents parsed with \olmocr have medium to high alignment with GPT-4o.
Increasing temperature unsurprisingly leads to a wider distribution of alignment scores, as noted by the increase of low matches for $\tau = 0.8$.

\begin{table}[!h]
    \vskip 0.15in
    \centering
    \begin{tabular}{cccc}
        \toprule
        \textbf{Model} & \textbf{Temperature}~$\tau$ & \textbf{Alignment}\\
        \midrule
        \textit{GPT-4o (self-alignment)} & \textit{0.1} & \textit{0.954}  \\
        GPT-4o mini & 0.1 & 0.833 \\
        \pipeline{} & 0.8 & 0.859 \\
        \pipeline{} & 0.1 & {\bf{0.875}} \\
        \bottomrule
    \end{tabular}
    \caption{Page-weighted alignment between GPT-4o, GPT-4o mini, and our fine-tuned model. We find that \model is more consistent with respect to its teacher than GPT-4o mini. Note that GPT-4o does not achieves a perfect alignment against itself due to the probabilistic nature of autoregressive decoding.}
    \label{tab:eval_alignment}
\end{table}

\begin{table}[!h]
    \centering
    \begin{tabular}{lrrr}
    \toprule
    \textbf{Name} & \textbf{Low match} & \textbf{Medium match} & \textbf{High match} \\
    \midrule
    \textit{GPT-4o (self alignment)} & \textit{38} & \textit{218} & \textit{965} \\
    GPT-4o mini & 214 & 478 & 529 \\
    \pipeline{} ($\tau=0.1$) & 158 & 363 & 700 \\
    \pipeline{} ($\tau=0.8$) & 195 & 390 & 636 \\
    \bottomrule
    \end{tabular}
    \caption{Match-up between olmOCR and different models compared to the \protect\train{} dataset. Low match indicates < 70\% alignment, Medium match is 70-95\% alignment, High match is >95\% alignment.}
    \label{tab:eval_matching}
\end{table}

\subsection{Intrinsic Human Evaluation}
\label{sec:elorating}
\paragraph{Experimental setup} To compare \pipeline{} against other common OCR methods, we collected pairwise human judgments of plain text produced by the three top ML-based PDF linearization tools---Marker, MinerU, and GOT-OCR 2.0---and calculating ELO ratings.

To create our evaluation set, we sample 2,017 new PDFs from the same distribution as used to create \train{} and run each PDF through \pipeline{} and the linearization tools mentioned above.
All other linearization tools were installed from either PyPI or Github according to their publicly available instructions as of January 14th, 2025. GOT-OCR 2.0 was configured in `format' mode, but otherwise all comparisons were done against default settings.

We then sampled 2,000 comparison pairs (same PDF, different tool).
We asked 11 data researchers and engineers at Ai2 to assess which output was the higher quality representation of the original PDF, focusing on reading order, comprehensiveness of content and representation of structured information. The user interface used is similar to that in Figure~\ref{fig:eval-template}. Exact participant instructions are listed in Appendix~\ref{sec:eloappendix}.

\paragraph{Evaluation results}

\begin{figure}[!h]
    \centering
    \includegraphics[width=0.45\linewidth]{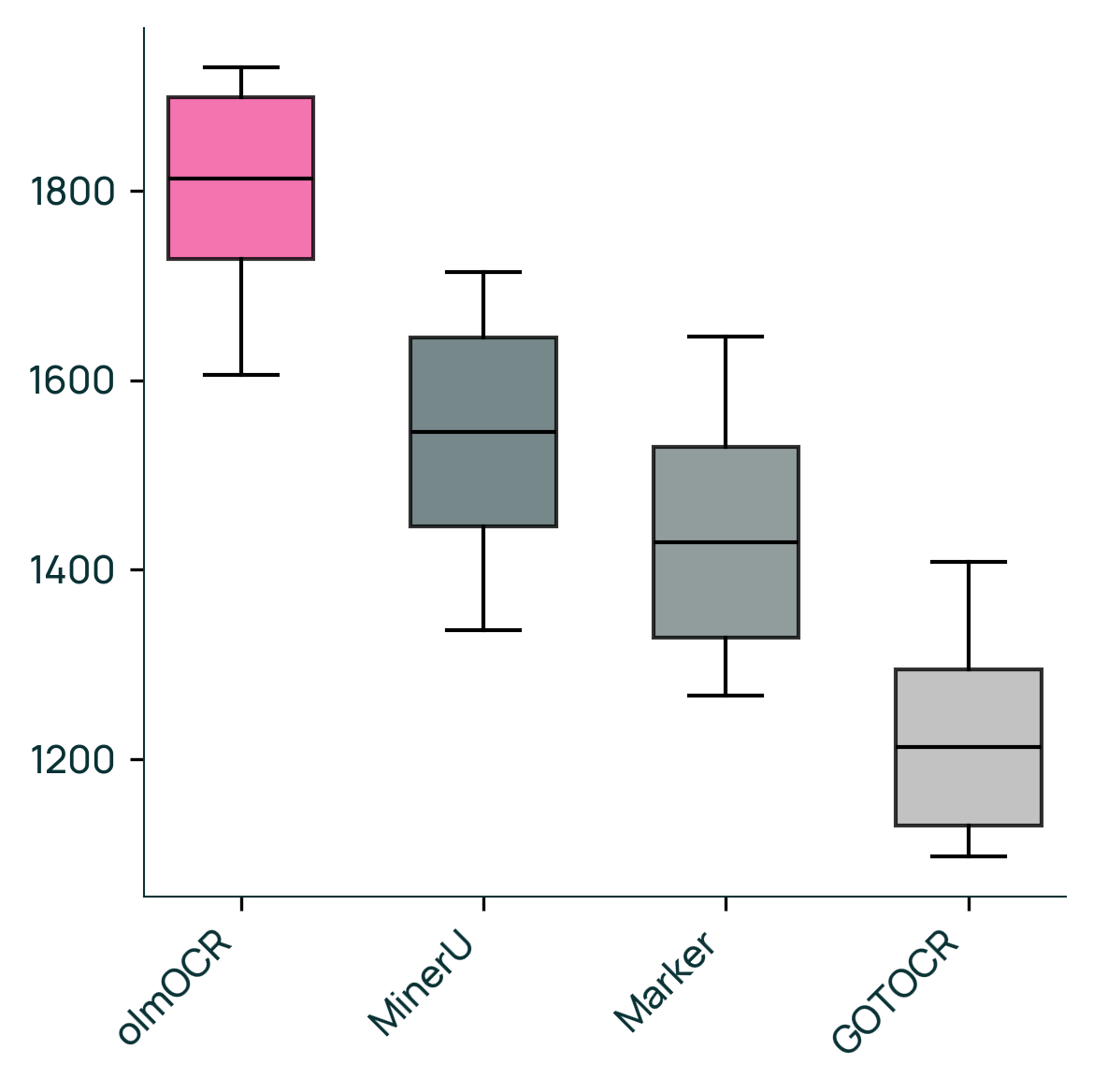}
    \caption{ELO ranking of \pipeline{} vs other popular PDF content extraction tools.}
    \label{fig:eloranking}
\end{figure}

We collected a total of 452 judgments where a participant expressed a preference between two models (the remaining 1,548 pairs were either skipped for being too similar, or marked as invalid).
On average, this is 75 judgments per pair of tools.
We calculate ELO ratings starting from a base of 1500 and report the average of 100 simulations to avoid ordering effects in ELO calculations; for 95\% confidence intervals, we use bootstrapping with 5000 resamples.

We visualize our results in Figure~\ref{fig:eloranking}. \pipeline{} achieves an ELO score over 1800, far exceeding all other PDF linearization tools.


\subsection{ELO Evaluation Instructions}
\label{sec:eloappendix}
In Section~\ref{sec:elorating}, we asked participants to compare the output of various common OCR tools
against \pipeline{}. Participants were given the instructions below, and presented with a document page, and the output of two random tools. They could then select which output was better, or select
`Both Good', `Both Bad', or `Invalid PDF' any of which would not count the comparison in the ELO ranking.

\subsubsection*{Instructions to participants}

\begin{lstlisting}[breaklines=true, frame=single, basicstyle=\ttfamily\footnotesize, keywordstyle=\bfseries\color{blue}, showstringspaces=false]
Compare the text in the two fields, and select which one better represents the contents of the document.

REMINDER: This is not about "the most faithful OCR", but "this OCR output seems really useful for training LMs"

- Does the text capture all of the meaningful content in the document in a natural order?
- Are the words correct (no weird incorrect words or split words)
- Is the whitepsace sensical?
- Is the useless header/footer content removed?
- Do the tables/equations look okay?

There is not a strict preference between Markdown and LaTeX, most importantly you should evaluate it on the text content, not which method was used to format it.

If you are not sure, or the document is in a language other than english, you can skip that entry, or mark "both good" "both bad", "invalid pdf".
\end{lstlisting}

\subsubsection*{ELO data}
We compute pairwise win/loss statistics between models to estimate relative performance under head-to-head comparisons. As shown in Table~\ref{tab:pairwise-stats}, \pipeline{} consistently outperforms other models such as \marker{}, \gotocr{}, and \mineru{}, with the highest win rate of 71.4\% against \mineru{}.

\begin{table}[htbp]
    \caption{Pairwise Win/Loss Statistics Between Models}
    \label{tab:pairwise-stats}

    \centering
    \begin{tabular}{lcc}
        \toprule
        Model Pair & Wins & Win Rate (\%) \\
        \midrule
        \pipeline{} vs. \marker{} & 49/31 & \textbf{61.3} \\
        \pipeline{} vs. \gotocr{} & 41/29 & \textbf{58.6} \\
        \pipeline{} vs. \mineru{} & 55/22 & \textbf{71.4} \\

        \marker{} vs. \mineru{} & 53/26 & 67.1 \\
        \marker{} vs. \gotocr{} & 45/26 & 63.4 \\
        \gotocr{} vs. \mineru{} & 38/37 & 50.7 \\

        \midrule
        Total & 452 & \\
        \bottomrule
    \end{tabular}

\end{table}

\section{Deploying \pipeline}
\subsection{Inference Pipeline}
To efficiently convert millions of documents, we develop the \olmocr pipeline using SGLang~\citep{zheng2024sglang} as the inference engine.
The pipeline batches documents into work items of around 500 pages each.
Each work item is then queued to run on a worker with access to a GPU for inference.
Optionally, workers can coordinate using a shared cloud bucket\footnote{We use Amazon Simple Storage Service (S3), but other cloud providers or network storage solutions could be easily used.}, allowing for batch jobs that scale from single nodes to hundreds of nodes without the need for complicated queue management.

We summarize our efforts by comparing operational costs of \olmocr against other API and local models in Table \ref{tab:inference_cost}.
Overall, we find \olmocr to be significantly more efficient than other pipelines.
It is over 32 times cheaper than GPT-4o in batch mode;
compared to other purposed-built pipelines and models, \olmocr is over 6 times cheaper than MinerU, and $1/3^{rd}$ of the cost of marker.

To balance maintaining high GPU utilization while also ensuring work items are completed
quickly, each worker queues up inference for all PDF pages in a work item simultaneously, and then
waits until the SGLang server has no more pending requests before proceeding to another work item in the queue.

\subsection{Increasing Robustness}

We implement several heuristics to improve reliability of \olmocr without compromising its throughput.

\paragraph{Prompt format}

During inference, we use the same abbreviated prompt described in Section~\S\ref{sec:model:training}.
This keeps the test time examples looking the same as what the model was trained on.
If the additional tokens generated by \method{} cause the overall prompt to exceed 8,192 tokens,
then we continue regenerating the \method{} tokens with exponentially lower character limits
until the overall prompt is of acceptable length.


\paragraph{Retries}
Unlike when we created \train{}, we do not enforce a specific JSON schema during inference on our fine-tuned model.
This is for two reasons:
first, we find that open source tools designed to force decode a sequence into a particular schema are unreliable, and that enforcing a schema which is even slightly off from what the model expects can cause generations to go out-of-domain or collapse into repetitions.
Second, and most importantly, we note that, since the model was extensively fine-tuned on the structured output,
it reliably adheres to the required schema without constraints.
For the rare cases when JSON parsing fails, we simply retry generating from the same input sequence.

\paragraph{Rotations}

The output JSON schema includes fields for \texttt{is\_rotation\_valid}
and \texttt{rotation\_correction}.
During inference, \pipeline{} pipeline reads these two fields and if \texttt{is\_rotation\_valid} is set to \texttt{true} it will rotate the page by the amount specified in \texttt{rotation\_correction} and reprocess the page.

\paragraph{Decoding}

In developing \olmocr, the most common failure we experience is outputs degenerating into endless repetitions of the same token, line, or paragraph.
This failure is caught automatically when the model's output either exceeds the maximum context length, or does not validate against our JSON schema.
We find that increasing generation temperature from $\tau=0.1$ up to $\tau=0.8$ reduces the likelihood of repetitions occurring.
Further, we modify \olmocr to reprocess failed pages up to N times, falling back to a plain text-based PDF extraction if the pipeline repeatedly fails.
This last mitigation is aided by the fact that \method{} randomly samples which anchors to include in the prompt;
thus, resampling can sometimes help the page process correctly by removing potentially problematic meta tokens.

We note that one one limitation of this approach is that, if retries occur often, the total generation throughput could be significantly reduced.
Further, letting generations repeat up to maximum sequence length uses significant memory within SGLang.
In future work, we plan to detect repeated generations sooner than at the maximum context length limit, and abort promptly.



\section{\train and \model Prompts }
\label{sec:gpt-4o-prompts}

\subsection{\train{} construction prompt for GPT-4o}
\label{sec:silver_prompt}

The prompt below was used to create the silver dataset, which we refer to as \train{} throughout the paper.
This dataset consists of structured outputs generated by GPT-4o, using images of PDF pages along with additional layout-aware textual features produced by our \method{} pipeline. We use this synthetic data to fine-tune our model.

In this prompt, the placeholder \texttt{\{base\_text\}} is replaced with the structured layout-aware text extracted from the PDF using \method{}.
The prompt instructs GPT-4o to output the natural reading-order text of the page, while respecting document semantics, suppressing hallucinations, and formatting content like equations and tables appropriately.

\begin{lstlisting}[breaklines=true, frame=single, basicstyle=\ttfamily\footnotesize, keywordstyle=\bfseries\color{blue}, showstringspaces=false]
Below is the image of one page of a PDF document, as well as some raw textual content that was previously extracted for it that includes position information for each image and block of text (The origin [0x0] of the coordinates is in the lower left corner of the image).
Just return the plain text representation of this document as if you were reading it naturally.
Turn equations into a LaTeX representation, and tables into markdown format. Remove the headers and footers, but keep references and footnotes.
Read any natural handwriting.
This is likely one page out of several in the document, so be sure to preserve any sentences that come from the previous page, or continue onto the next page, exactly as they are.
If there is no text at all that you think you should read, you can output null.
Do not hallucinate.
RAW_TEXT_START
{base_text}
RAW_TEXT_END
\end{lstlisting}

\subsubsection*{JSON Schema used to prompt GPT-4o}
\label{sec:json_schema}

\begin{lstlisting}[breaklines=true, frame=single, basicstyle=\ttfamily\footnotesize, keywordstyle=\bfseries\color{blue}, showstringspaces=false]
"json_schema": {
            "name": "page_response",
            "schema": {
                "type": "object",
                "properties": {
                    "primary_language": {
                        "type": ["string", "null"],
                        "description": "The primary language of the text using two-letter codes or null if there is no text at all that you think you should read.",
                    },
                    "is_rotation_valid": {
                        "type": "boolean",
                        "description": "Is this page oriented correctly for reading? Answer only considering the textual content, do not factor in the rotation of any charts, tables, drawings, or figures.",
                    },
                    "rotation_correction": {
                        "type": "integer",
                        "description": "Indicates the degree of clockwise rotation needed if the page is not oriented correctly.",
                        "enum": [0, 90, 180, 270],
                        "default": 0,
                    },
                    "is_table": {
                        "type": "boolean",
                        "description": "Indicates if the majority of the page content is in tabular format.",
                    },
                    "is_diagram": {
                        "type": "boolean",
                        "description": "Indicates if the majority of the page content is a visual diagram.",
                    },
                    "natural_text": {
                        "type": ["string", "null"],
                        "description": "The natural text content extracted from the page.",
                    },
                },
                "additionalProperties": False,
                "required": [
                    "primary_language",
                    "is_rotation_valid",
                    "rotation_correction",
                    "is_table",
                    "is_diagram",
                    "natural_text",
                ],
            },
            "strict": True,
        },
\end{lstlisting}

\subsection{\model{} prompt}
\label{sec:fine_tune_prompt}

The prompt below is used to draw responses from our fine-tuned model during inference. As before, the placeholder \texttt{\{base\_text\}} is replaced with the output of the \method{} pipeline i.e., layout-aware textual features extracted from the PDF page.

\begin{lstlisting}[breaklines=true, frame=single, basicstyle=\ttfamily\footnotesize, keywordstyle=\bfseries\color{blue}, showstringspaces=false]
Below is the image of one page of a document, as well as some raw textual content that was previously extracted for it.
Just return the plain text representation of this document as if you were reading it naturally.
Do not hallucinate.
RAW_TEXT_START
{base_text}
RAW_TEXT_END
\end{lstlisting}

\subsection{\train{} Classification Prompt}
\label{sec:classification_prompt}

The prompt and structured schema below was used to classify a sample of documents from \train{} as reported
in Table~\ref{tab:pdf_types}.

\begin{lstlisting}[breaklines=true, frame=single, basicstyle=\ttfamily\footnotesize, keywordstyle=\bfseries\color{blue}, showstringspaces=false]
This is an image of a document page, please classify it into one of the following categories that best overall summarizes its nature: academic, legal, brochure, slideshow, table, diagram, or other. Also determine the primary language of the document and your confidence in the classification (0-1).
\end{lstlisting}

\begin{lstlisting}[breaklines=true, frame=single, basicstyle=\ttfamily\footnotesize, keywordstyle=\bfseries\color{blue}, showstringspaces=false]
class DocumentCategory(str, Enum):
    ACADEMIC = "academic"
    LEGAL = "legal"
    BROCHURE = "brochure"
    SLIDESHOW = "slideshow"
    TABLE = "table"
    DIAGRAM = "diagram"
    OTHER = "other"

class DocumentClassification(BaseModel):
    category: DocumentCategory
    language: str
    confidence: float
\end{lstlisting}

\subsection{\train PII Prompt}
\label{sec:train-set-pii}

We implemented comprehensive prompting for detecting personally identifiable information (PII) in the documents while cleaning the \train:
\begin{lstlisting}[breaklines=true, frame=single, basicstyle=\ttfamily\footnotesize, keywordstyle=\bfseries\color{blue}, showstringspaces=false]
You are a document analyzer that identifies Personally Identifiable Information
(PII) in documents.
Your task is to analyze the provided document image and determine:
1. Whether the document is intended for public release or dissemination
   (e.g., research paper, public report, etc.)
2. If the document contains any PII

For PII identification, follow these specific guidelines:
IDENTIFIERS FOR PII:
The following are considered identifiers that can make information PII:
- Names (full names, first names, last names, nicknames)
- Email addresses
- Phone numbers

PII THAT MUST CO-OCCUR WITH AN IDENTIFIER:
The following types of information should ONLY be marked as PII if they occur
ALONGSIDE an identifier (commonly, a person's name):
- Addresses (street address, postal code, etc.)
- Biographical Information (date of birth, place of birth, gender, sexual
  orientation, race, ethnicity, citizenship/immigration status, religion)
- Location Information (geolocations, specific coordinates)
- Employment Information (job titles, workplace names, employment history)
- Education Information (school names, degrees, transcripts)
- Medical Information (health records, diagnoses, genetic or neural data)

PII THAT OCCURS EVEN WITHOUT AN IDENTIFIER:
The following should ALWAYS be marked as PII even if they do not occur
alongside an identifier:
- Government IDs (Social Security Numbers, passport numbers, driver's license
  numbers, tax IDs)
- Financial Information (credit card numbers, bank account/routing numbers)
- Biometric Data (fingerprints, retina scans, facial recognition data,
  voice signatures)
- Login information (ONLY mark as PII when a username, password, and login
  location are present together)

If the document is a form, then only consider fields which are filled out
with specific values as potential PII.
If this page does not itself contain PII, but references documents
(such as curriculum vitae, personal statements) that typically contain PII,
then do not mark it as PII.
Only consider actual occurrences of the PII within the document shown.
\end{lstlisting}




\newpage
\clearpage

\section{Further details of \bench}
\label{appendix:bench01}

\subsection{Data sources}

See Table~\ref{tab:benchmark_composition} for further details about PDFs selected for each \bench{} category, the source, and the processing.

\begin{table}[ht]
\caption{Document source category breakdown of \bench}
\centering
\begin{tabular}{lcccc}
\toprule
\textbf{Category} & \textbf{PDFs} & \textbf{Tests} & \textbf{Source} & \textbf{Extraction Method} \\
\midrule
arXiv\_math        & 522 & 2,927 & arXiv & Dynamic programming alignment \\
old\_scans\_math   & 36  & 458   & Internet Archive & Script-generated + manual rules \\
tables\_tests      & 188 & 1,020 & Internal repository & \texttt{gemini-flash-2.0} \\
old\_scans         & 98  & 526   & Library of Congress & Manual rules \\
headers\_footers   & 266 & 753   & Internal repository & DocLayout-YOLO + \texttt{gemini-flash-2.0} \\
multi\_column      & 231 & 884   & Internal repository & \texttt{claude-sonnet-3.7} + HTML rendering \\
long\_tiny\_text   & 62  & 442   & Internet Archive & \texttt{gemini-flash-2.0} \\
\midrule
\textbf{Total}     & 1,403 & 7,010 & Multiple sources & \\
\bottomrule
\end{tabular}
\label{tab:benchmark_composition}
\end{table}

\subsection{Prompting Strategies and Implementation Details}
This section provides comprehensive documentation of the prompting techniques and design strategies to make \bench. These prompting approaches were critical in generating test cases while utilizing LLMs and ensuring consistency across document categories.
\subsubsection{Mathematical Expressions}
For generating mathematical expression test cases from old scans, we employed direct prompts focused on precision. This concise prompt architecture proved effective in extracting LaTeX representations minimizing hallucination. The explicit instruction to use standard LaTeX delimiters (\$\$) ensured consistent formatting across the \bench.
\begin{lstlisting}[breaklines=true, frame=single, basicstyle=\ttfamily\footnotesize, keywordstyle=\bfseries\color{blue}, showstringspaces=false]
Please extract the mathematical equations from the document without
omission. Always output the mathematical equations as Latex escaped
with $$. Do not hallucinate.
\end{lstlisting}

\subsubsection{Multi-column}
For Multi-column documents, we utilized a two-stage prompting strategy. The initial analytical stage established structural context:

\begin{lstlisting}[breaklines=true, frame=single, basicstyle=\ttfamily\footnotesize, keywordstyle=\bfseries\color{blue}, showstringspaces=false]
Analyze this document and provide a detailed assessment of its structure.
Focus on the layout, headings, footers, and any complex formatting.
Please be precise.
\end{lstlisting}

This preliminary analysis was incorporated into a subsequent HTML rendering prompt:

\begin{lstlisting}[breaklines=true, frame=single, basicstyle=\ttfamily\footnotesize, keywordstyle=\bfseries\color{blue}, showstringspaces=false]
Render this document as clean, semantic HTML. Here is the analysis of the
document structure:

{analysis_text}

Requirements:
1. Use appropriate HTML tags for headings, paragraphs, and lists.
2. Use <header> and <footer> for top and bottom content.
3. For images, use a placeholder <div> with class 'image'.
4. Render math equations inline using \( \) or \[ \].
5. Preserve any multi-column layout using CSS flexbox or grid.
6. The viewport is fixed at {png_width // 2}x{png_height // 2} pixels.

Enclose your HTML in a html code block.
\end{lstlisting}

This approach significantly helped in layout preservation in complex documents by providing explicit dimensional constraints and structural information.

\label{sec:pii}
\subsubsection{PII Detection and Filtering}
We use the same PII detection and filtering as for construction \train{}; see Appendix~\ref{sec:train-set-pii}.

\subsubsection{Cleaning Mathematical Expressions}
Mathematical expression verification employed specialized prompting for validating equation presence and accuracy:
\begin{lstlisting}[breaklines=true, frame=single, basicstyle=\ttfamily\footnotesize, keywordstyle=\bfseries\color{blue}, showstringspaces=false]
This is a mathematical expression verification task.
I'm showing you a page from a PDF document containing mathematical expressions.
Please verify if the following LaTeX expression:
{latex_expression}
appears correctly in the document.
Respond with a JSON object containing:
1. "status": "correct" or "incorrect"
2. "confidence": a value between 0 and 1 representing your confidence in the answer
3. "explanation": a brief explanation of why you believe the expression is correct or incorrect
Focus specifically on checking if this exact mathematical expression appears in the document.
\end{lstlisting}

\subsubsection{Cleaning Reading Order Tests}
For natural reading order test cases, we implemented below verification prompt to ensure appropriate text segment relationships:

\begin{lstlisting}[breaklines=true, frame=single, basicstyle=\ttfamily\footnotesize, keywordstyle=\bfseries\color{blue}, showstringspaces=false]
Does the text in the 'before' field and the 'after' field appear in the same region of the page?
Look at the PDF image and determine if these texts are located near each other or in completely
different parts of the page. Different regions could be the captions for different images, or
inside of different insets or tables. However, appearing the same column of text, or in the
naturally flowing next column of text is close enough.

Before: {before_text}

After: {after_text}

Respond with 'YES' if they appear in the same region or column, and 'NO' if they appear in
different regions. Then explain your reasoning in 1-2 sentences.
\end{lstlisting}

\subsubsection{Header and Footer Verification}

For validating header and footer text identification, we employed JSON-structured verification prompts:

\begin{lstlisting}[breaklines=true, frame=single, basicstyle=\ttfamily\footnotesize, keywordstyle=\bfseries\color{blue}, showstringspaces=false]
This is a header and footer verification task.
I'm showing you a page from a PDF document containing headers and footers text.
Please verify if the headers or footers is exactly matches the below text.
{header_footer_text}
Respond with a JSON object containing:
1. "status": "correct" or "incorrect"
2. "confidence": a value between 0 and 1 representing your confidence in the answer
3. "explanation": a brief explanation of why you believe the text is correct or incorrect
Focus specifically on checking if this exact header or footer expression appears in the document.
\end{lstlisting}

Our prompting strategy deliberately requested different output formats for different content types (Markdown for general text, LaTeX for equations, HTML for tables) to optimize representation fidelity across diverse document elements. Low temperature settings (typically 0.1) was maintained across all the prompt executions to ensure reproducible outputs, particularly important for establishing consistent test cases.

\subsection{Sample Test Classes}
\label{sec:samples}
Below are are few examples taken from \bench \\

\begin{figure}[h]
    \centering
    \includegraphics[height=12cm]{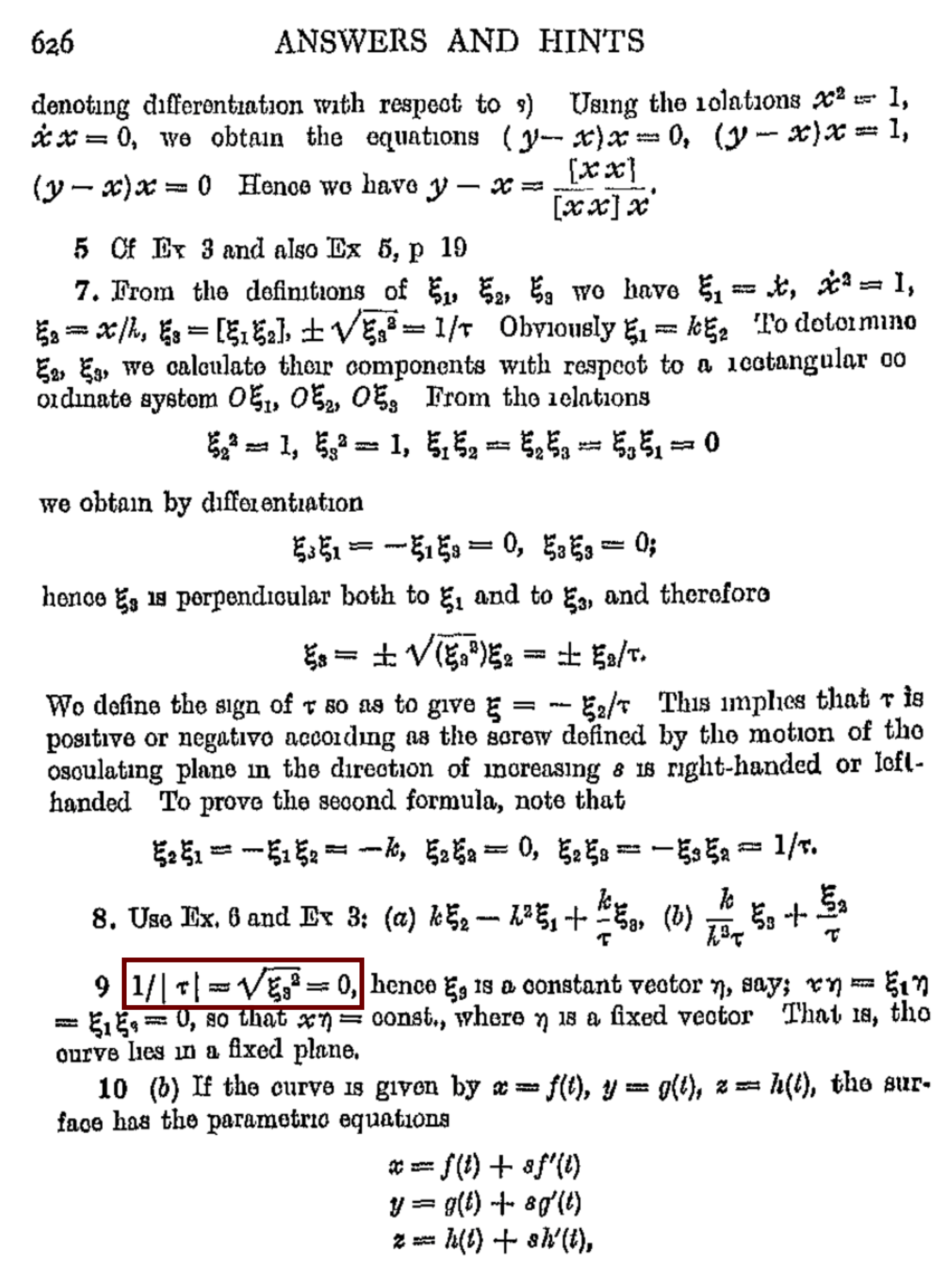}
    \caption{Sample visualization from old\_scans\_math. The OCR output for the highlighted equation should be: \texttt{1/|$\backslash$tau| = $\backslash$sqrt\{$\backslash$xi\_\{3\}\^{}\{2\}\} = 0}}
    \label{fig:old_scans_math_example}
\end{figure}
\vspace{.5cm}
\begin{figure}[h]
    \centering
    \includegraphics[height=8cm]{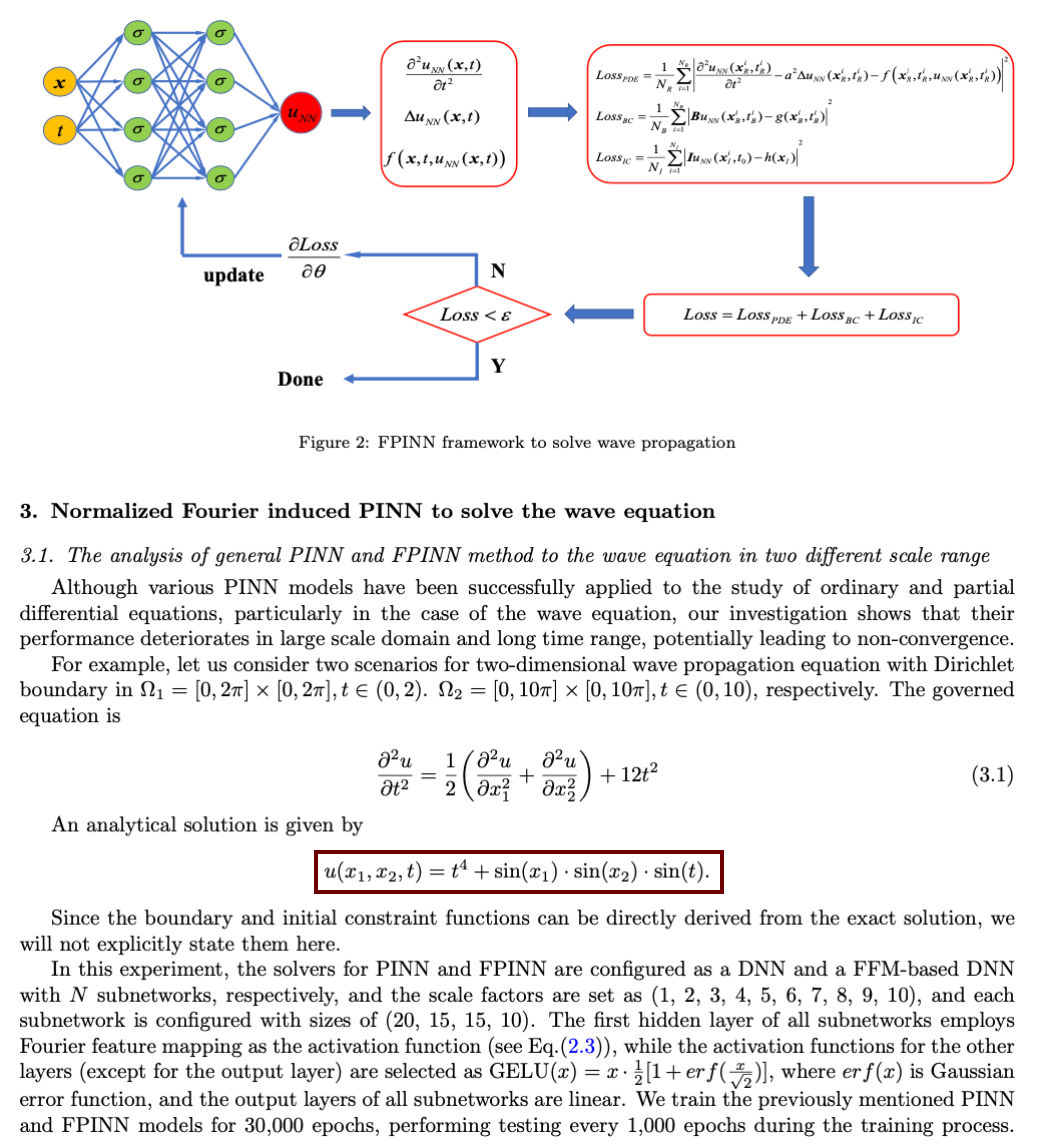}
    \caption{Sample visualization of a math equation from arXiv\_math. The OCR output for the highlighted equation should be: \texttt{u(x\_\{1\},x\_\{2\},t)=t\^{}\{4\} + $\backslash$text\{sin\}(x\_\{1\}) $\backslash$cdot $\backslash$text\{sin\}(x\_\{2\}) $\backslash$cdot $\backslash$text\{sin\}(t)}}
    \label{fig:arxiv_math_example}
\end{figure}
\vspace{1.5cm}
\begin{figure}[h]
    \centering
    \includegraphics[height=11cm]{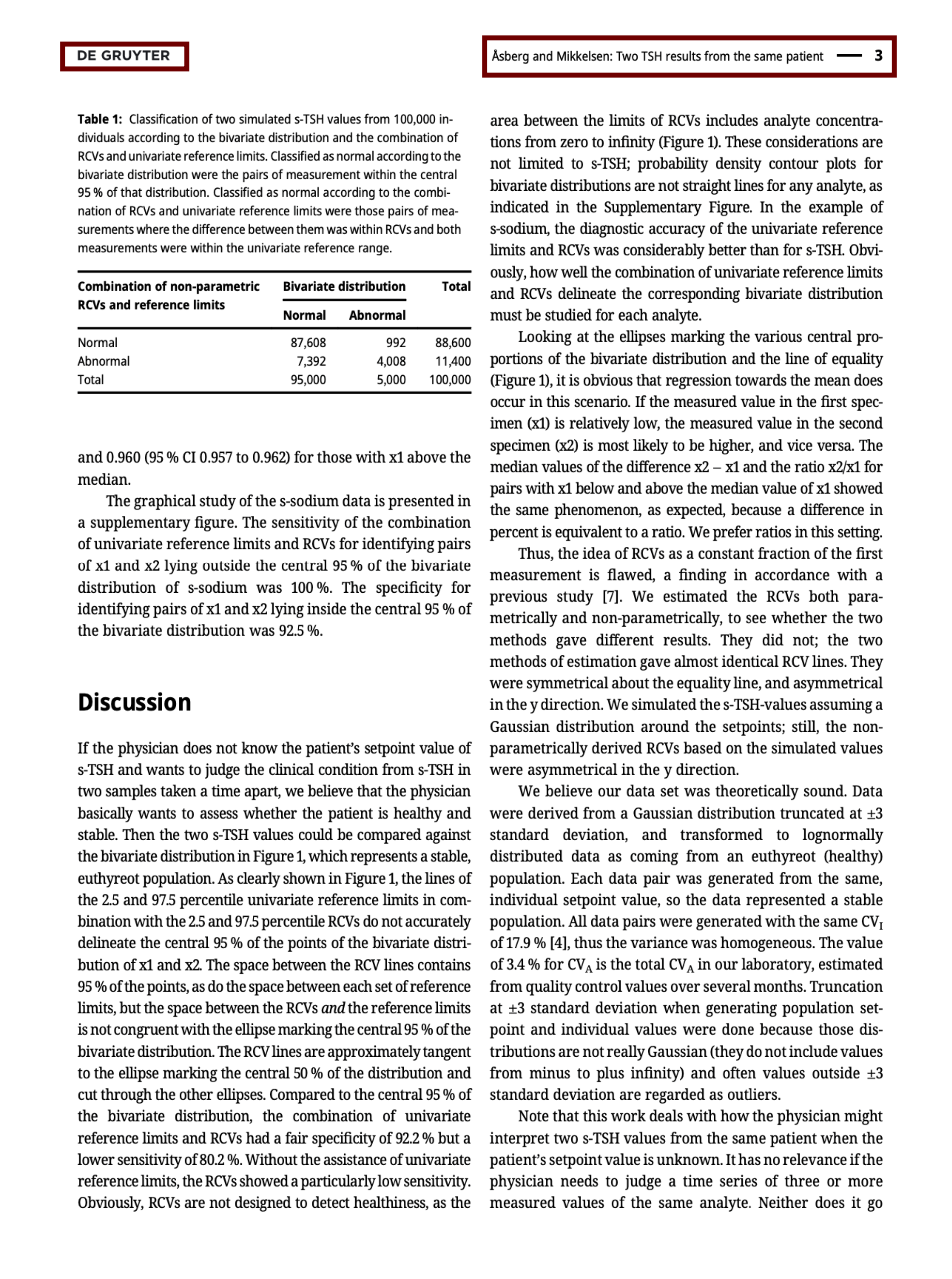}
    \caption{Sample visualization of \texttt{headers\_footers}. We want the OCR to skip the document headers and page number.}
    \label{fig:headers_footers_example}
\end{figure}
\begin{figure}[h]
    \centering
    \includegraphics[height=7cm]{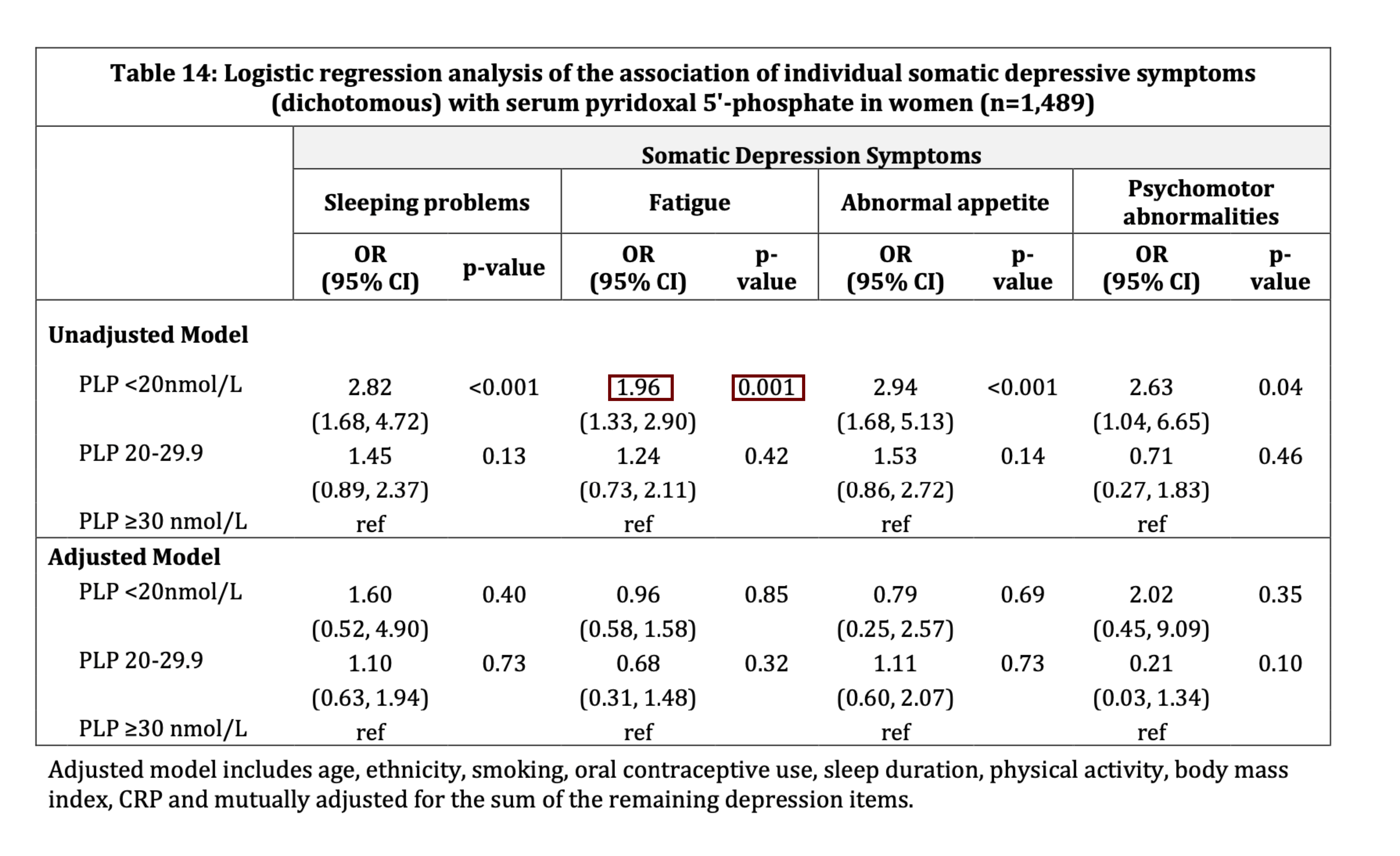}
    \caption{Sample visualization of \texttt{table\_tests}. We want the OCR to predict that cell 1.96 is to the left of cell 0.001.}
    \label{fig:table_tests_example}
\end{figure}
\begin{figure}[h]
    \centering
    \includegraphics[height=11cm]{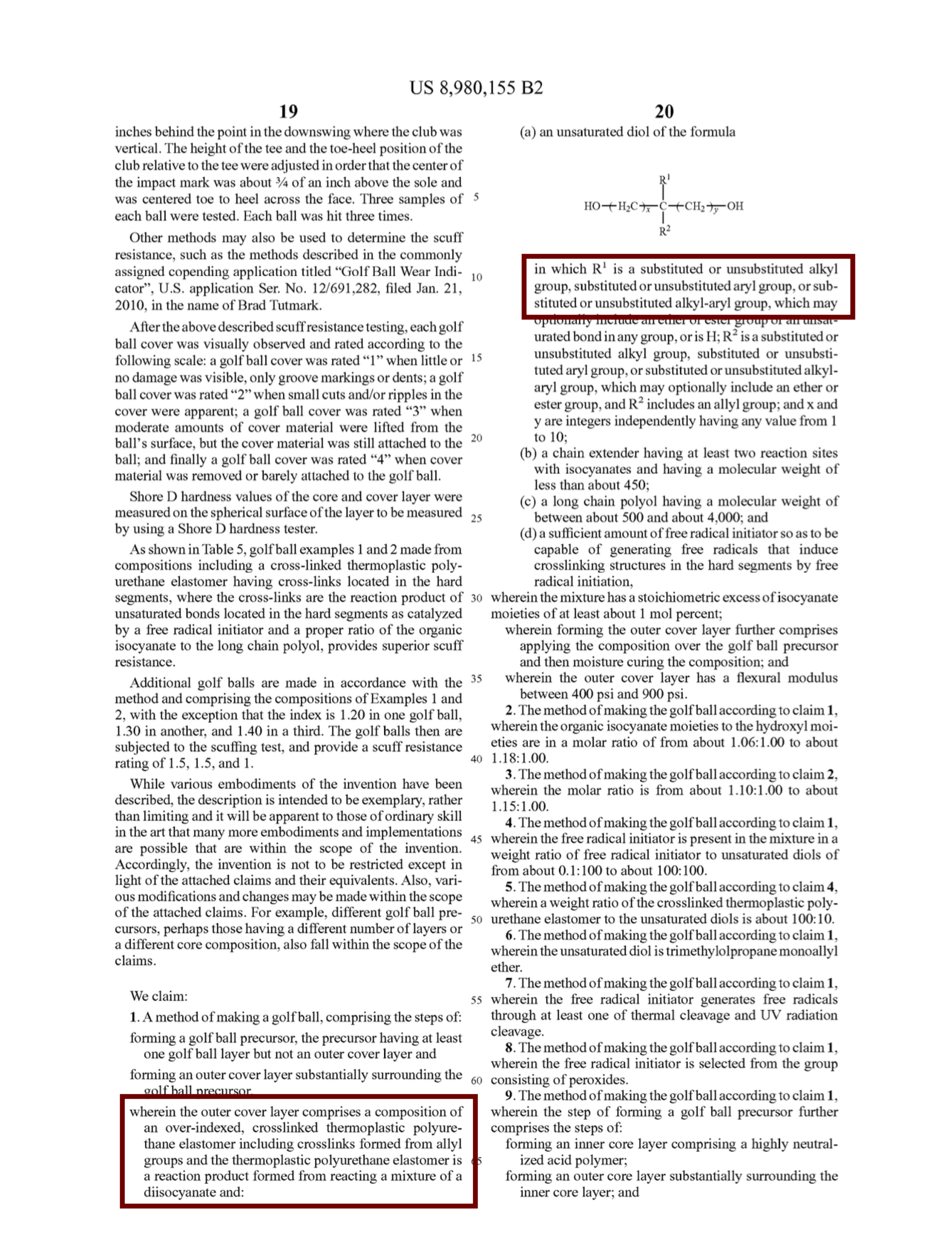}
    \caption{Sample visualization of \texttt{reading\_order}. The reading order should start with the left column before moving to the right column.}
    \label{fig:reading_order_example}
\end{figure}

\newpage
\clearpage

\newpage
\clearpage

\newpage
\section{Example output}
\label{appendix:olmocr_output}
Below are some sample outputs on particularly challenging data. \pipeline{}, MinerU, GOT-OCR 2.0 and Marker run with default settings.

\begin{table}[h!]
  \centering
  \small
  \begin{tabular}{p{3.6cm}p{3.6cm}p{3.6cm}p{3.6cm}}
    \multicolumn{4}{c}{
      \begin{tabular}{@{}l@{}}
        \includegraphics[height=8cm]{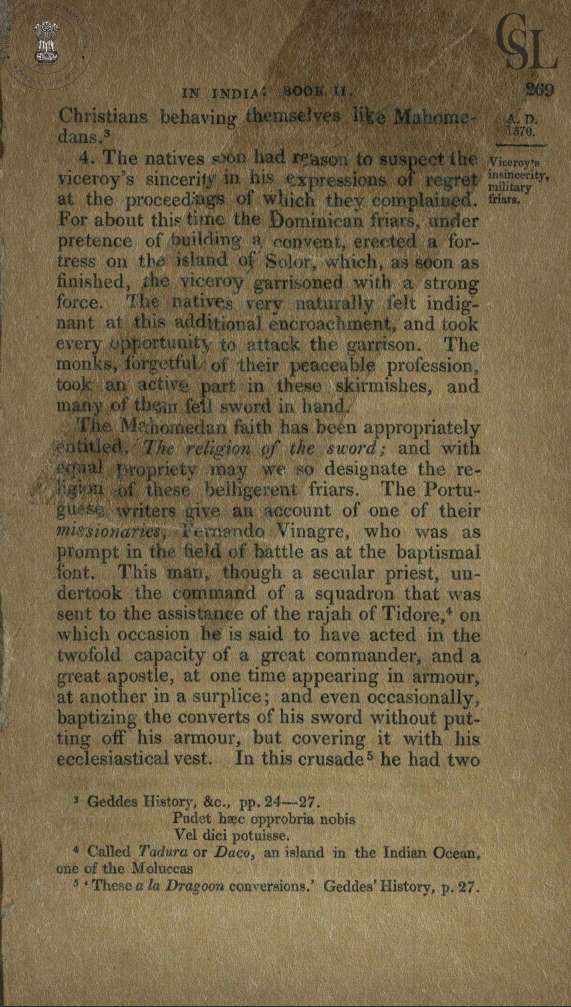}
      \end{tabular}
    } \\
    \textbf{\pipeline} & \textbf{MinerU} & \textbf{GOT-OCR 2.0} & \textbf{Marker} \\
    \midrule
        \tiny
        Christians behaving themselves like Mahomedans.

        4. The natives soon had reason to suspect the viceroy's sincerity in his expressions of regret at the proceedings of which they complained. For about this time the Dominican friars, under pretence of building a convent, erected a fortress on the island of Solor, which, as soon as finished, the viceroy garrisoned with a strong force. The natives very naturally felt indignant at this additional encroachment, and took every opportunity to attack the garrison. The monks, forgetful of their peaceable profession, took an active part in these skirmishes, and many of them fell sword in hand.

        The Mahomedan faith has been appropriately entitled, The religion of the sword; and with equal propriety may we so designate the religion of these belligerent friars. The Portuguese writers give an account of one of their missionaries, Fernando Vinagre, who was as prompt in the field of battle as at the baptismal font. This man, though a secular priest, undertook the command of a squadron that was sent to the assistance of the rajah of Tidore, on which occasion he is said to have acted in the twofold capacity of a great commander, and a great apostle, at one time appearing in armour, at another in a surplice; and even occasionally, baptizing the converts of his sword without putting off his armour, but covering it with his ecclesiastical vest. In this crusade he had two
    &
  \tiny
      ININDIASY BOOKU
Christians bchaving.themselves like Mahome dans.3

4.The natives soon had reason to suspect ihe viceroy's sincerity in his expressions of regret at the proceedings of which they complained. For about this time the Dominican friars,under pretenceof building a convent,erected a for tress on the island of Solorwhich,as soon as finishedthe viceroy garrisoned with a strong force. The natives very naturally felt indig nant at this additional encroachment, and took every pportunity to attack the garrison.The monks,forgetful of their peaceable profession took an activa part in these skirmishes, and many of tbein feil sword in hand.

TheMahornedan faithhas been appropriately ntitled.The religion of the swordand with equal propriety may we so designate the region of these belligerent friars.The Portugueswriters give an account of one of their missionarzes,femando Vinagre,who was as prompt in the field of battle as at the baptismal font. This man, though a secular priest, undertook the command of a squadron that was sent to the assistance of the rajah of Tidore,4 on which occasion he is said to have acted in the twofold capacity of a great commander, and a great apostle, at one time appearing in armour, at another in a surplice;and even occasionally baptizing the converts of his sword without put ting off his armour, but covering it with his ecclesiastical vest.In this crusadehe had two
   &
\tiny
      IN INDIA:  BOOK U 269 Christians behaving themselves like Mahome-  1670.  4. The natives son had reason to suspect the Viceroy' s vice roy' s sincerity in his expressions of regret in s in e eri ty at the proceedings of which they complained.  fl it ars.  For about this time the Dominican f mars, under pre ten ce of building a convent, erected a for-  tress on the island of Sol or, which, as soon as finished, the vice roy garrisoned with a strong force. The natives very naturally felt indig-  nant at this additional encroachment, and took every opportunity to attack the garrison. The monks, forgetful of their peaceable profession,  took an active part in these skirmishes, and many of the n fell sword in hand.  The Mh on med an faith has been appropriately entitled. The religion of the sword; and with e ral Tropriety may we so designate the re-  gian of these belligerent friars. The Port u-  gue s writers give an account of one of their mission are s, Fer endo Vina gre, who was as prompt in the fe ld of battle as at the baptismal font. This man, though a secular priest, un-  der took the command of a squadron that was sent to the assistance of the rajah of Tidore, on which occasion he is said to have acted in the twofold capacity of a great commander, and a great apostle, at one time appearing in armour,  at another in a surplice; and even occasionally,  baptizing the converts of his sword without put-  ting off his armour, but covering it with his ecclesiastical vest. In this crusade he had two 3 Ged des History, \& c. , pp. 24-27.  P ude th aec opp rob ria nobis Vel die ipo tui sse.  Called Tadur u or Daco, an island in the Indian Ocean,  one of the Mol ucc as These a laDra goon conversions. Ged des History, p. 27.
    &
    \tiny
\#\# **IN INDIA *** BOOK TI. S69
Christians behaving themselves like Ma borne- a. dans.3 .\textquotedblleft5/0-
*t>.*

The natives soon had reason to suspect the viceroy, viceroy's sincerity in his expressions of regret at the proceedings of which they complained. \textquotedblleft n.\textquotedblleft' For about this time the Dominican friars, under pretence of building a. convent, erected a fortress on the island of Sol or, which, as soon as finished, the viceroy garrisoned with a strong force. The natives' very naturally felt indig-S nant at this additional encroachment, and took every opportunity to attack the garrison. The monks, forgetful/ of their peaceable profession, took an active part in these skirmishes, and many of tbg.tr fell sword in hand.

The i'lfinomedan faith has been appropriately entitled., \textquoteleft The religion of the sword\textquoteright,; and with equal propriety may we so designate the re- .\ i'gv.m of these belligerent friars. The Portugu writers give an account of one of their \textquoteleft missionaries,\textquoteright\ Fernando Vinagre, who was as prompt in the field of battle as at the baptismal font. This man, though a secular priest, undertook the command of a squadron that was I sent to the assistance of the rajah of Tidore,4 on which occasion he is said to have acted in the twofold capacity of a great commander, and a great apostle, at one time appearing in armour, ; at another in a surplice; and even occasionally, baptizing the converts of his sword without putting off his armour, but covering it with his ecclesiastical vest. In this crusade5 he had two
\> 3 Geddes History, \&c., pp. 24---27. Pudet h\ae c opprobria nobis Vel dici potuisse.
\> 4 Called \textquoteleft T a d u ra\textquoteright\ or \textquoteleft D a c o,\textquoteright\ an island in the Indian Ocean, one of the Moluccas
\> 5 \textquoteleft These \textquoteleft a la D ra g o o n\textquoteright\ conversions.\textquoteright\ Geddes' History, p. 27.
  \\
    \bottomrule
  \end{tabular}
\end{table}

\newpage

\begin{table}[h!]
  \centering
  \small
  \begin{tabular}{p{3.6cm}p{3.6cm}p{3.6cm}p{3.6cm}}
    \multicolumn{4}{c}{
      \begin{tabular}{@{}l@{}}
        \includegraphics[height=7cm]{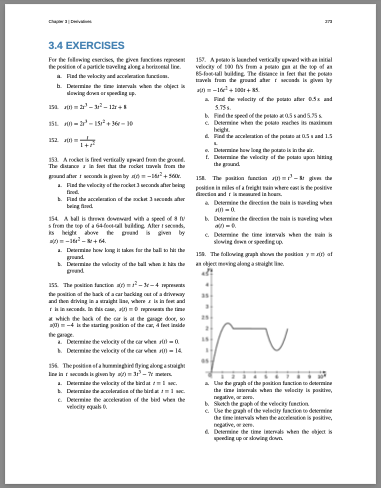}
      \end{tabular}
    } \\
    \textbf{\pipeline} & \textbf{MinerU} & \textbf{GOT-OCR 2.0} & \textbf{Marker} \\
    \midrule
        \tiny
3.4 EXERCISES

For the following exercises, the given functions represent the position of a particle traveling along a horizontal line.

a. Find the velocity and acceleration functions.

b. Determine the time intervals when the object is slowing down or speeding up.

150. \( s(t) = 2t^3 - 3t^2 - 12t + 8 \)

151. \( s(t) = 2t^3 - 15t^2 + 36t - 10 \)

152. \( s(t) = \frac{t}{1 + t^2} \)

153. A rocket is fired vertically upward from the ground. The distance \( s \) in feet that the rocket travels from the ground after \( t \) seconds is given by \( s(t) = -16t^2 + 560t \).

a. Find the velocity of the rocket 3 seconds after being fired.

b. Find the acceleration of the rocket 3 seconds after being fired.

154. A ball is thrown downward with a speed of 8 ft/s from the top of a 64-foot-tall building. After \( t \) seconds, its height above the ground is given by \( s(t) = -16t^2 - 8t + 64 \).

a. Determine how long it takes for the ball to hit the ground.

b. Determine the velocity of the ball when it hits the ground.

155. The position function \( s(t) = t^2 - 3t - 4 \) represents the position of the back of a car backing out of a driveway and then driving in a straight line, where \( s \) is in feet and \( t \) is in seconds. In this case, \( s(t) = 0 \) represents the time at which the back of the car is at the garage door, so \( s(0) = -4 \) is the starting position of the car, 4 feet inside the garage.

a. Determine the velocity of the car when \( s(t) = 0 \).

b. Determine the velocity of the car when \( s(t) = 14 \).

156. The position of a hummingbird flying along a straight line in \( t \) seconds is given by \( s(t) = 3t^3 - 7t \) meters.

a. Determine the velocity of the bird at \( t = 1 \) sec.

b. Determine the acceleration of the bird at \( t = 1 \) sec.

c. Determine the acceleration of the bird when the velocity equals 0.

157. A potato is launched vertically upward with an initial velocity of 100 ft/s from a potato gun at the top of an 85-foot-tall building. The distance in feet that the potato travels from the ground after \( t \) seconds is given by \( s(t) = -16t^2 + 100t + 85 \).
\ldots
    &
  \tiny
\# 3.4 EXERCISES

For the following exercises, the given functions represent the position of a particle traveling along a horizontal line.

a. Find the velocity and acceleration functions. b. Determine the time intervals when the object is slowing down or speeding up.

150. $s(t)=2t^{3}-3t^{2}-12t+8$
151. $s(t)=2t^{3}-15t^{2}+36t-10$
152. $s(t)=\frac{t}{1+t^{2}}$

153. A rocket is fired vertically upward from the ground. The distance $s$ in feet that the rocket travels from the ground after $t$ seconds is given by $s(t)=-16t^{2}+560t,$ .

a. Find the velocity of the rocket 3 seconds after being fired.
b. Find the acceleration of the rocket 3 seconds after being fired.

154. A ball is thrown downward with a speed of 8 ft/ s from the top of a 64-foot-tall building. After $t$ seconds, its height above the ground is given by $s(t)=-16t^{2}-8t+64.$ .

a. Determine how long it takes for the ball to hit the ground.
b. Determine the velocity of the ball when it hits the ground.

155. The position function $s(t)=t^{2}-3t-4$ represents the position of the back of a car backing out of a driveway and then driving in a straight line, where $s$ is in feet and $t$ is in seconds. In this case, $s(t)=0$ represents the time at which the back of the car is at the garage door, so $s(0)=-4$ is the starting position of the car, 4 feet inside the garage.

a. Determine the velocity of the car when $s(t)=0$ .
b. Determine the velocity of the car when $s(t)=14$ .

156. The position of a hummingbird flying along a straight line in $t$ seconds is given by $s(t)=3t^{3}-7t$ meters.

a. Determine the velocity of the bird at $t=1$ sec. b. Determine the acceleration of the bird at $t=1$ sec. c. Determine the acceleration of the bird when the velocity equals 0.

157. A potato is launched vertically upward with an initial velocity of 100 ft/s from a potato gun at the top of an 85-foot-tall building. The distance in feet that the potato travels from the ground after $t$ seconds is given by $s(t)=-16t^{2}+100t+85.$ .

\ldots
   &
\tiny
Chapter 3 | Derivatives
273
3.4 EXERCISES
For the following exercises, the given functions represent
the position of a particle traveling along a horizontal line.
a.
Find the velocity and acceleration functions.
b.
Determine the time intervals when the object is
slowing down or speeding up.
150.
s(t) = 2t3 −3t2 −12t + 8
151.
s(t) = 2t3 −15t2 + 36t −10
152.
s(t) =
t
1 + t2
153.
A rocket is ﬁred vertically upward from the ground.
The distance
s
in
feet
that
the
rocket
travels
from
the
ground after
t seconds is given by
s(t) = −16t2 + 560t.
a.
Find the velocity of the rocket 3 seconds after being
ﬁred.
b.
Find the acceleration of the rocket 3 seconds after
being ﬁred.
154.
A ball is thrown downward with a speed of 8 ft/
s from the top of a 64-foot-tall building. After t seconds,
its
height
above
the
ground
is
given
by
s(t) = −16t2 −8t + 64.
a.
Determine how long it takes for the ball to hit the
ground.
b.
Determine the velocity of the ball when it hits the
ground.
155.
The position function
s(t) = t2 −3t −4 represents
the position of the back of a car backing out of a driveway
and then driving in a straight line, where
s
is in feet and
t is in seconds. In this case, s(t) = 0 represents the time
at which the back of the car is at the garage door, so
s(0) = −4 is the starting position of the car, 4 feet inside
the garage.
a.
Determine the velocity of the car when
s(t) = 0.
b.
Determine the velocity of the car when
s(t) = 14.
156.
The position of a hummingbird ﬂying along a straight
line in
t seconds is given by
s(t) = 3t3 −7t
2
2
2
2
2
2
2
2
2
2
3
3
3
3
3
3
3
3
3
3
4
4
4
4
4
4
4
4
4
4
5
5
5
5
5
5
5
5
5
5
4
4
4
4
4
4
4
4
4
3
3
3
3
3
3
3
3
3
1
1
1
1
1
1
1
1
1
1
3
3
3
3
3
3
3
3
3
0
0
0
0
0
0
0
0
0
0
1
1
1
1
1
1
1
1
1
2
2
2
2
2
2
2
2
2
1
1
1
1
1
1
1
1
1
0
0
0
0
0
0
0
0
0
3
3
3
3
3
3
3
3
3
2
2
2
2
2
2
2
2
2
4
4
4
4
4
4
4
4
4
2
2
2
2
2
2
2
2
2
0
1
1
1
1
1
1
1
1
3
4
4
4
4
4
4
4
4
3
4
4
4
4
4
4
4
4
2
3
3
3
3
3
3
3
3
0
1
1
1
1
1
1
1
1
0
1
1
1
1
1
1
1
1
2
1
1
1
1
1
1
1
1
0
2
2
2
2
2
2
2
2
2
5
5
5
5
5
5
5
5
5
1
1
1
1
1
1
1
1
1
5
5
5
5
5
5
5
5
5
0
0
0
0
0
0
0
0
0
5
5
5
5
5
5
5
5
5
3
3
3
3
3
3
3
3
3
5
5
5
5
5
5
5
5
5
2
2
2
2
2
2
2
2
2
a.
Use the graph of the position function to determine
the time intervals when the velocity is positive,
negative, or zero.
b.
Sketch the graph of the velocity function.
c.
Use the graph of the velocity function to determine
the time intervals when the acceleration is positive,
negative, or zero.
d.
Determine the time intervals when the object is
speeding up or slowing down.
\ldots
    &
    \tiny
\#\# **3.4 EXERCISES**

For the following exercises, the given functions represent the position of a particle traveling along a horizontal line.

- a. Find the velocity and acceleration functions.
- b. Determine the time intervals when the object is slowing down or speeding up.

$$150. \quad s(t) = 2t^3 - 3t^2 - 12t + 8$$

$$151. \quad s(t) = 2t^3 - 15t^2 + 36t - 10t$$

$$152. \quad s(t) = \frac{t}{1+t^2}$$

153. A rocket is fired vertically upward from the ground. The distance *s* in feet that the rocket travels from the ground after *t* seconds is given by *s*(*t*) = −16*t* 2 + 560*t*.

- a. Find the velocity of the rocket 3 seconds after being fired.
- b. Find the acceleration of the rocket 3 seconds after being fired.

154. A ball is thrown downward with a speed of 8 ft/ s from the top of a 64-foot-tall building. After *t* seconds, its height above the ground is given by *s*(*t*) = −16*t* 2 − 8*t* + 64.

- a. Determine how long it takes for the ball to hit the ground.
- b. Determine the velocity of the ball when it hits the ground.

155. The position function *s*(*t*) = *t* 2 − 3*t* − 4 represents the position of the back of a car backing out of a driveway and then driving in a straight line, where *s* is in feet and *t* is in seconds. In this case, *s*(*t*) = 0 represents the time at which the back of the car is at the garage door, so *s*(0) = −4 is the starting position of the car, 4 feet inside the garage.

- a. Determine the velocity of the car when *s*(*t*) = 0.
- b. Determine the velocity of the car when *s*(*t*) = 14.

156. The position of a hummingbird flying along a straight line in *t* seconds is given by *s*(*t*) = 3*t* 3 − 7*t* meters.

- a. Determine the velocity of the bird at *t* = 1 sec.
- b. Determine the acceleration of the bird at *t* = 1 sec.
- c. Determine the acceleration of the bird when the velocity equals 0.

\ldots
  \\
    \bottomrule
  \end{tabular}
\end{table}

\newpage

\begin{table}[h!]
  \centering
  \small
  \begin{tabular}{p{3.6cm}p{3.6cm}p{3.6cm}p{3.6cm}}
    \multicolumn{4}{c}{
      \begin{tabular}{@{}l@{}}
        \includegraphics[height=8cm]{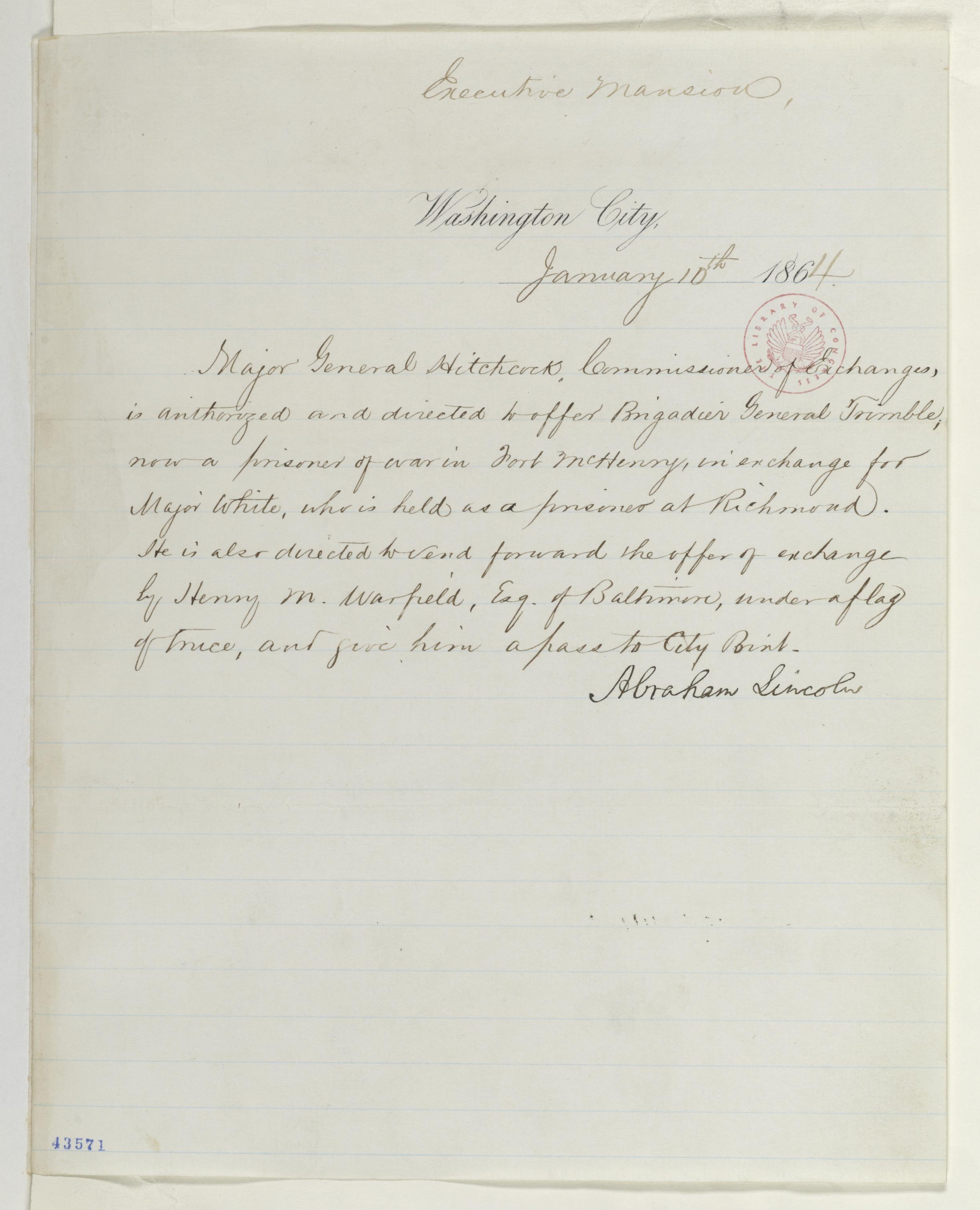}
      \end{tabular}
    } \\
    \textbf{\pipeline} & \textbf{MinerU} & \textbf{GOT-OCR 2.0} & \textbf{Marker} \\
    \midrule
        \small
Executive Mansion,

Washington City,

January 15th, 1864

Major General Hitchcock, Commissioner of Exchanges, is authorized and directed to offer Brigadier General Trimble, now a prisoner of war in Fort McHenry, in exchange for Major White, who is held as a prisoner at Richmond. He is also directed to send forward the offer of exchange by Henry M. Warfield, Esq. of Baltimore, under a flag of truce, and give him a pass to City Point.

Abraham Lincoln
    &
  \small
\textit{No text produced.}
   &
\small
43571
Bachington City
January 10th 1864.
Major General Architect, Commissioner of aivachangera
is authorized and directed by ffeed Bngader General Trelmble,
new a firemen of war in Fert nchery in exchange for
Mayor White, who held a a firemen at Hillmannd.
He is aker conducted by end forward the offer of exchange
by Henry in. Warfield, Lag. of Balthmore, under a flag
of three, and five him afaies to City Bink.
Abraham Lincoln
    &
    \small
necuhve Mansion Vastington amany layor Seneral Hitchcocks Commissioner of Cachanges, is anthonged and directed to offer Bingadier General Trin prisoner of war in Fort Inctienny, in exchange now w Major White, who is held as a preises at Richmond Ite is also directed to vand forwards the offer of exchange by Stenny in. Warfield, Eag. of Baltimore, under aflag 11 mice, and give him apass to tity Point. Abrakan Sincolus
  \\
    \bottomrule
  \end{tabular}
\end{table}

\end{document}